\documentclass[]{spie}

\makeatletter
\def\endthebibliography{%
  \def\@noitemerr{\@latex@warning{Empty `thebibliography' environment}}%
  \endlist
}
\makeatother

\usepackage{amsmath,amsfonts,amssymb}
\usepackage{graphicx}
\usepackage[colorlinks=true, allcolors=blue]{hyperref}
\usepackage{caption}
\usepackage{subcaption}
\usepackage{float}
\usepackage{algorithm,algorithmicx}

\newcommand{\etal}{\textit{et al.}}
\newcommand{\ie}{\textit{i.e.}}
\newcommand{\foc}{\mathfrak{f}}
\DeclareMathOperator*{\argmin}{arg\,min}

\makeatletter
\def\BState{\State\hskip-\ALG@thistlm}
\makeatother

\graphicspath{{./figures/}}

\title{Investigation of moving objects through atmospheric turbulence from a non-stationary platform}

\author[ab]{Nicholas Ferrante}
\author[b]{J\'er\^ome Gilles}
\author[a]{Shibin Parameswaran}
\affil[a]{Naval Information Warfare Center Pacific, 53560 Hull St, San Diego, CA, USA}
\affil[b]{San Diego State University, 5500 Campanile Dr, San Diego, CA, USA}

\authorinfo{Further author information:\\
Nicholas Ferrante: E-mail: nicholas.ferrante1@navy.mil, Telephone: +1 619-553-7877\\
J\'er\^ome Gilles: E-mail: jgilles@sdsu.edu, Telephone: +1 619-594-7240 \\
Shibin Parameswaran: E-mail: paramesw@spawar.navy.mil, Telephone: +1 619-553-1554}

\begin{document} 
\maketitle

\begin{abstract}
In this work, we extract the optical flow field corresponding to moving objects from an image sequence of a scene impacted by atmospheric turbulence \emph{and}  captured from a moving camera. 
Our procedure first computes the optical flow field and creates a motion model to compensate for the flow field induced by camera motion.
After subtracting the motion model from the optical flow, we proceed with our previous work, Gilles et al~\cite{gilles2018detection}, where a spatial-temporal cartoon+texture inspired decomposition is performed on the motion-compensated 
flow field in order to separate flows corresponding to atmospheric turbulence and object motion.
Finally, the geometric component is processed with the detection and tracking method and is compared against a ground truth.
All of the sequences and code used in this work are open source and are available by contacting the authors.
\end{abstract}


\section{INTRODUCTION}\label{chap:intro}

In image processing, the detection and tracking of objects in an image sequence is a classic problem.
Typically, the problem is broken into two steps: detection and tracking. 
The detection step has multiple variations in the literature: background subtraction \cite{shaikh2014moving}, temporal differencing 
\cite{yi2010moving}, optical flow \cite{chauhan2013moving}, and most recently the use of deep learning \cite{lecun2015deep}.
With the assumption of static camera, the method of background subtraction uses a running average of past frames to build and update a 
background model.
When a new frame is introduced, the difference of the current frame and the foreground is assumed to correspond to real world motion.
Optical flow based detection estimates a displacement vector field for each pixel in the image.
The magnitude of this vector field is computed and any value above a certain threshold is assumed to correspond to a moving object.
Neural networks exploit the relationship between object detection and image understanding in tracking high-level features within an image set.
Once detected, a tracking algorithm is utilized in order to investigate the behavior of detections.
The primary methods of tracking in image processing are point based tracking (\textit{e.g.} Kalman~\cite{patel2013moving} or particle filters~\cite{gordon2004beyond}), kernel based tracking (\textit{e.g.} Simple Template Matching~\cite{briechle2001template}, Mean shift~\cite{comaniciu2000real}, Support Vector Machine~\cite{avidan2001support}, Layer based tracking~\cite{zhou2003background}), and silhouette based tracking (contour and shape matching \cite{rosenhahn2006comparison}).
A comparison of these methods is presented by Balaji \cite{balaji2017survey}.

In the case of a moving camera, there is an added complication of motion introduced through rigid body motion.
In order to compensate for the false motion, a camera motion model is needed.
One of the primary methods for creating a motion model is to use image registration between frames in order to find a relation across feature points.
Feature points are correlated between frames and their displacements are then used to fit a model for camera motion.
With the camera motion model, any locations in the image that do not conform to the predicted location are presumed to be moving objects.
A review of object detection in video images captured by a moving camera is presented by Yaxdi \cite{yazdi2018new}.

In this work, we add an extra level of difficulty by considering the presence of atmospheric turbulence in our image sequence.
Atmospheric turbulence is typically due to an energy exchange within a fluid that separates an observer and an object.
In our case, this fluid is air that is impacted by a temperature gradient, driving the change in the density of air and thus the change in the index of refraction.
Through Snell's\cite{giancoli2008physics} law, the change in the index of refraction causes light to bend as it moves between a target and the observer \cite{hyde2010determining}.
This phenomenon introduces geometric deformation and blur within an image scene, bringing with it a myriad of problems in terms of image processing.
With the temporal variance of the index of refraction, random oscillatory movement is introduced within the image scene that causes the false appearance of motion and degrade feature points needed for a camera motion model, therefore motivating a need for an object detection algorithm that may handle the combined impact of atmospheric turbulence as well as camera motion.

In this work, we introduce an optical flow based method that builds a camera motion model that is effective regardless of the impact of atmospheric turbulence.
Once we build a camera motion model, we then are tasked with detecting moving objects within a camera motion compensated optical flow field.
The proposed solution to this problem of detection is motivated by our prior work, Gilles \etal~\cite{gilles2018detection}, where a preprocessing method was introduced with the assumption of a static camera.
In the proposed work, we extend the existing algorithm to detect and track objects from an optical flow field that contains atmospheric turbulence and global camera motion.

\subsection{Data Collection}\label{sec: Data Collection}

\begin{figure}[t] 
	\centering
    \hfill
   \begin{subfigure}[c]{.33\textwidth}
        \centering
        \includegraphics[width = .95\textwidth]{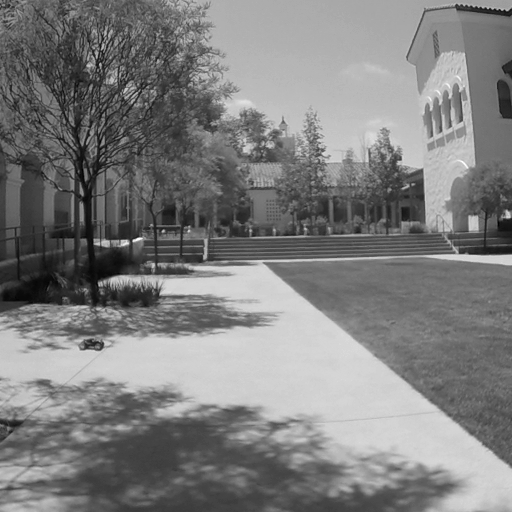}
    \end{subfigure}%
    \hfill
   \begin{subfigure}[c]{.33\textwidth}
        \centering
        \includegraphics[width = .95\textwidth]{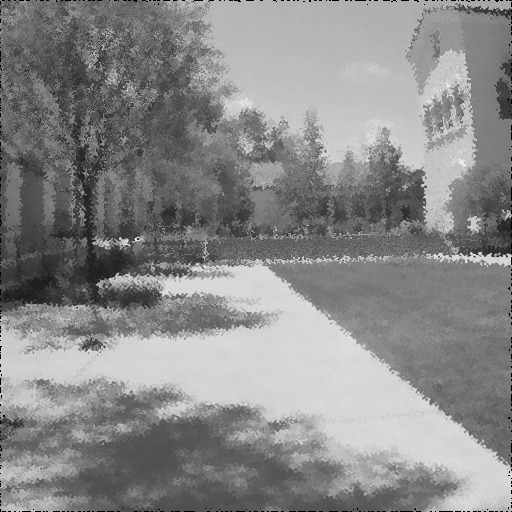}
    \end{subfigure}
    \hfill
   \begin{subfigure}[c]{.33\textwidth}
        \centering
        \includegraphics[width = .95\textwidth]{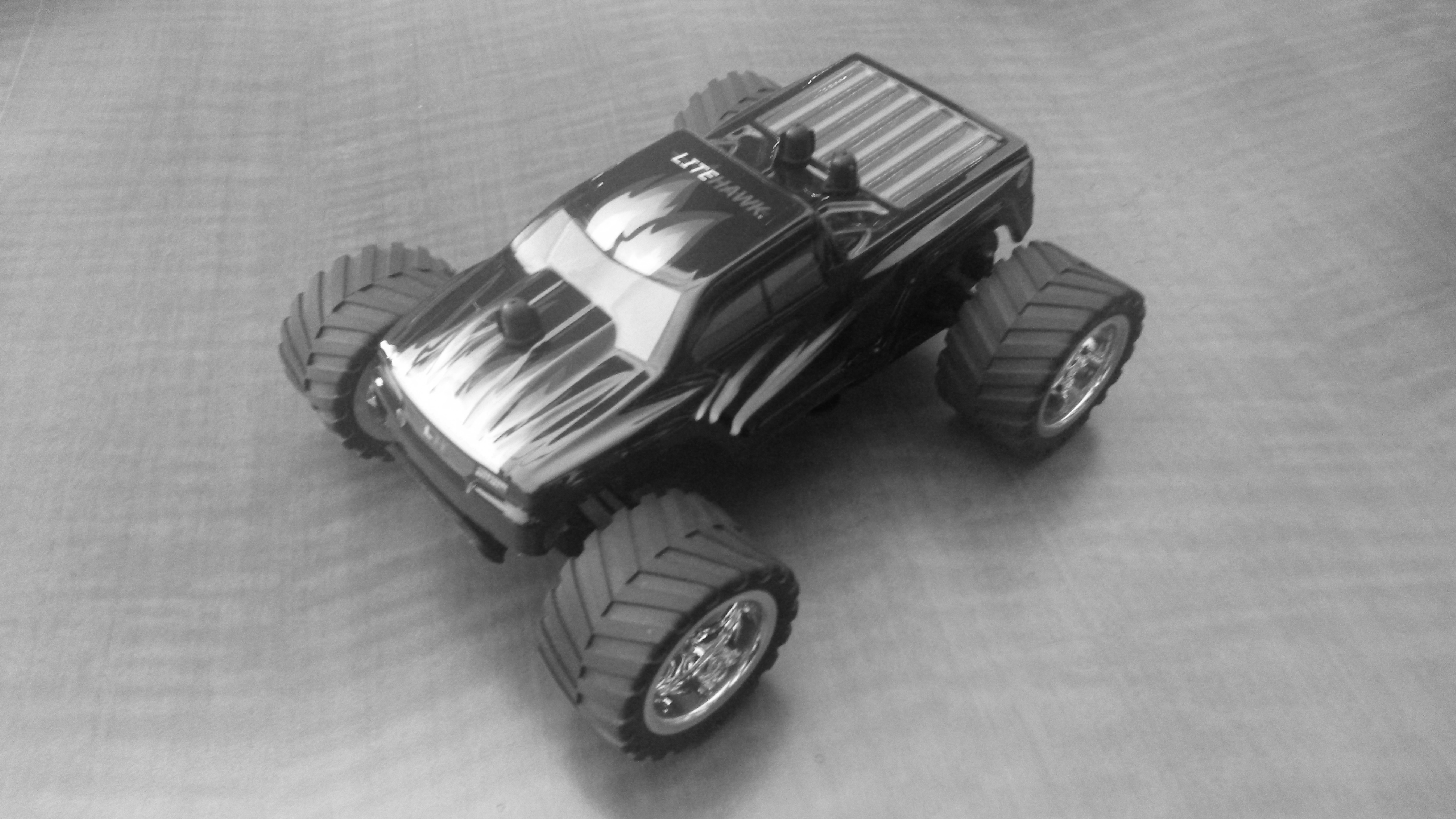}
    \end{subfigure}
    \hfill
 \caption{Frame 30 from the \textit{Courtyard4} sequence is shown (left) along with its corresponding version with the introduction of simulated atmospheric turbulence (center). On the right is a close view of the remote control car used in each sequence.}
  \label{fig: sim turbulence}
 \end{figure}

In order to properly test the methods proposed in this project, we needed a dataset that contains both atmospheric turbulence and global camera motion.
Unfortunately, before the start of this project, there were no data sources available that contained small moving objects along with camera motion and atmospheric turbulence.
Therefore, we created our own dataset to act as an addendum to the preexisting Open Turbulence Image Set (OTIS \footnote{\url{https://zenodo.org/communities/otis/}})~\cite{gilles2017open}.
In this section, we discuss the utilized equipment, methods of data processing, and some examples taken from the dataset.
All sequences were recorded with a GoPro Hero 4 Black camera modified with a RibCage Air chassis, permitting the use of several lens types.
A small tripod was used to hold the camera and it was manually rotated and translated in order to produce global camera motion.
In order to introduce a moving object in the image sequence, a remote control car was utilized (see Figure~\ref{fig: sim turbulence}).
Data sequences without atmospheric turbulence were captured in the Engineering and Interdisciplinary Sciences Complex courtyard at San Diego State University.
Data sequences with naturally forming atmospheric turbulence were taken at Peterson Gym 600 practice field, San Diego State University, during the afternoon hours on 24 July 2018.\\
\begin{table}[t]
\caption{Open source data set sequences from OTIS used in this research.}
 \label{tbl: OTIS}
\centering
 \begin{tabular}{||c|c|c||} 
 \hline
 Name & Size  & Turbulence\\
 \hline\hline 
 Courtyard1 & $[512 \times 512 \times 100]$  & Simulated \\ 
 Courtyard2 & $[350 \times 350 \times 113]$  & Simulated \\
 Courtyard3 & $[300 \times 300 \times 138]$  & Simulated \\
 Courtyard4 & $[512 \times 512 \times 100]$  & Simulated \\
 Field1 		& $[512 \times 512 \times 100]$  & Natural \\ 
 Field2 		& $[300 \times 300 \times 100]$ & Natural \\
 Field3 		& $[512 \times 512 \times 101]$ & Natural \\
 Field4 		& $[300 \times 300 \times 120]$ & Natural \\
 Field5 		& $[300 \times 300 \times 72]$ & Natural \\
 \hline
 \end{tabular}
\end{table}
Once the image sequences were captured, the MP4 files were downloaded onto a Macintosh computer, converted to PNG image files with the \textit{ffmpeg} command\footnote{\url{https://ffmpeg.org/}}, and cropped using the \textit{imagemagick} crop command\footnote{\url{http://www.imagemagick.org/}}  to contain a field of view with the moving object.
For the sequences containing atmospheric turbulence, no further processing was needed, but for the sequences that were taken in the courtyard, an atmospherc turbulence simulator, developed by Tahtali, Fraser, and Lambert \cite{tahtali2005restoration}, was implemented.
The result from one frame is shown in Figure~\ref{fig: sim turbulence}.
It is important to notice that around the edges of the simulated sequence there exist black regions that may cause added complications during object detection.
Therefore we recommend that the user crop the image edges in order to remove these artifacts.
A summary of the datasets used in this work is shown in Table~\ref{tbl: OTIS}. \\
When assessing the effectiveness of our method, we require a ground truth for each of the sequences.  Rather than manually observing the location of the car in each sequence, we coded a MATLAB application which provides a graphical user interface (GUI) to easily create a multiple-track ground truth which then can be used by the performance metric function.

\subsection{Optical Flow Computation}\label{section: Optical Flow}

\begin{figure}[b]
\begin{center}
   \begin{minipage}{2in}
        \centering
        \includegraphics[width = .9\textwidth]{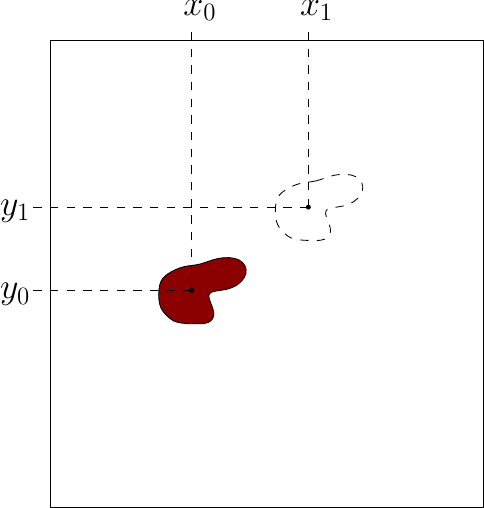}
    \end{minipage}
    \begin{minipage}{2in}
        \centering
        \includegraphics[width = .9\textwidth]{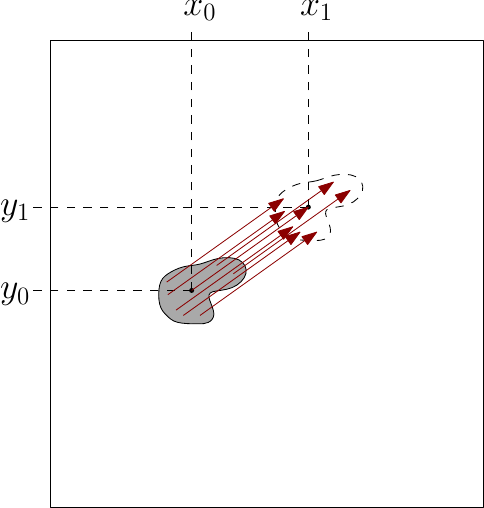}
    \end{minipage}
    \begin{minipage}{2in}
        \centering
        \includegraphics[width = .9\textwidth]{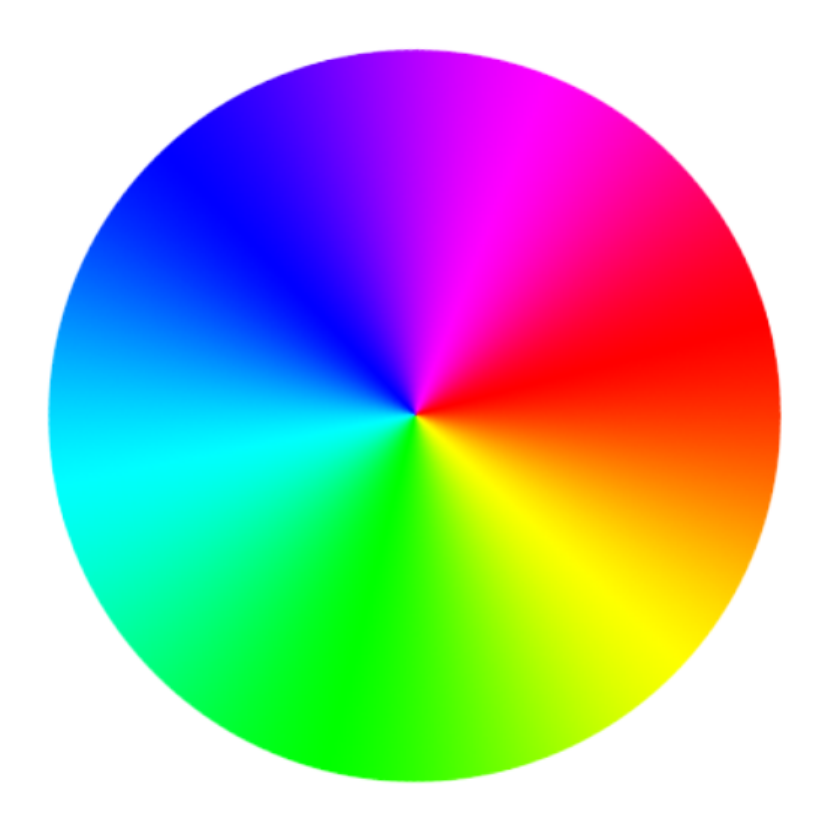}
    \end{minipage}
\end{center}
\caption{Illustration of the concept of optical flow. The left image represents an object (the red area) as its initial position centered around $(x_0,y_0)$ and its position in the next frame (the dashed area) centered around $(x_1,y_1)$. In the center figure, the set of arrows corresponds to the motion vectors \ie optical flow. The colorwheel on the right provides the correspondence between color and movement direction.}
\label{fig:Optical Flow Example 1}
\end{figure}
The nature of this work does not lend itself well to detecting objects in the image domain, therefore we focus our efforts in the optical flow domain.
The optical flow of an image sequence is the spatiotemporal vector field corresponding to the spatial movement of each pixel from one frame to the next.
A general example of what we refer to by optical flow is demonstrated in Figure~\ref{fig:Optical Flow Example 1} where we have a region that moves from coordinates centered about $(x_0,y_0)$ to new coordinates $(x_1,y_1)$.  As the optical flow field is dense within the image, we are unable to properly visualize the vectors in their traditional form, therefore we employ a technique which associates the direction of each vector with a color on the colorwheel from Figure~\ref{fig:Optical Flow Example 1} with the intensity of the color corresponding to the magnitude of the vector.
There are many ways that optical flow may be computed, but for the purposes of this work we will primarily use the Horn-Schunck (HS)~\cite{horn1981determining} and $TV-L^1$ algorithm~\cite{zach2007duality}.
These algorithms are both interfaced through MATLAB by using C code provided by the Image Processing OnLine (IPOL) journal.

\section{Global Camera Motion}
\begin{figure}[b]
\centering
\begin{subfigure}[b]{.1995\textwidth}
\centering
\includegraphics[width = .99\textwidth]{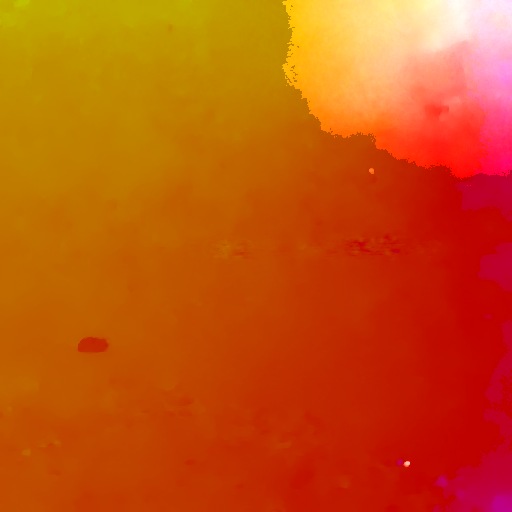} 
\caption{Colorized Flow}\label{fig:Camera Motion Motivation 1 a}
\end{subfigure} 
\begin{subfigure}[b]{.209\textwidth}
\centering
\includegraphics[width = .99\textwidth]{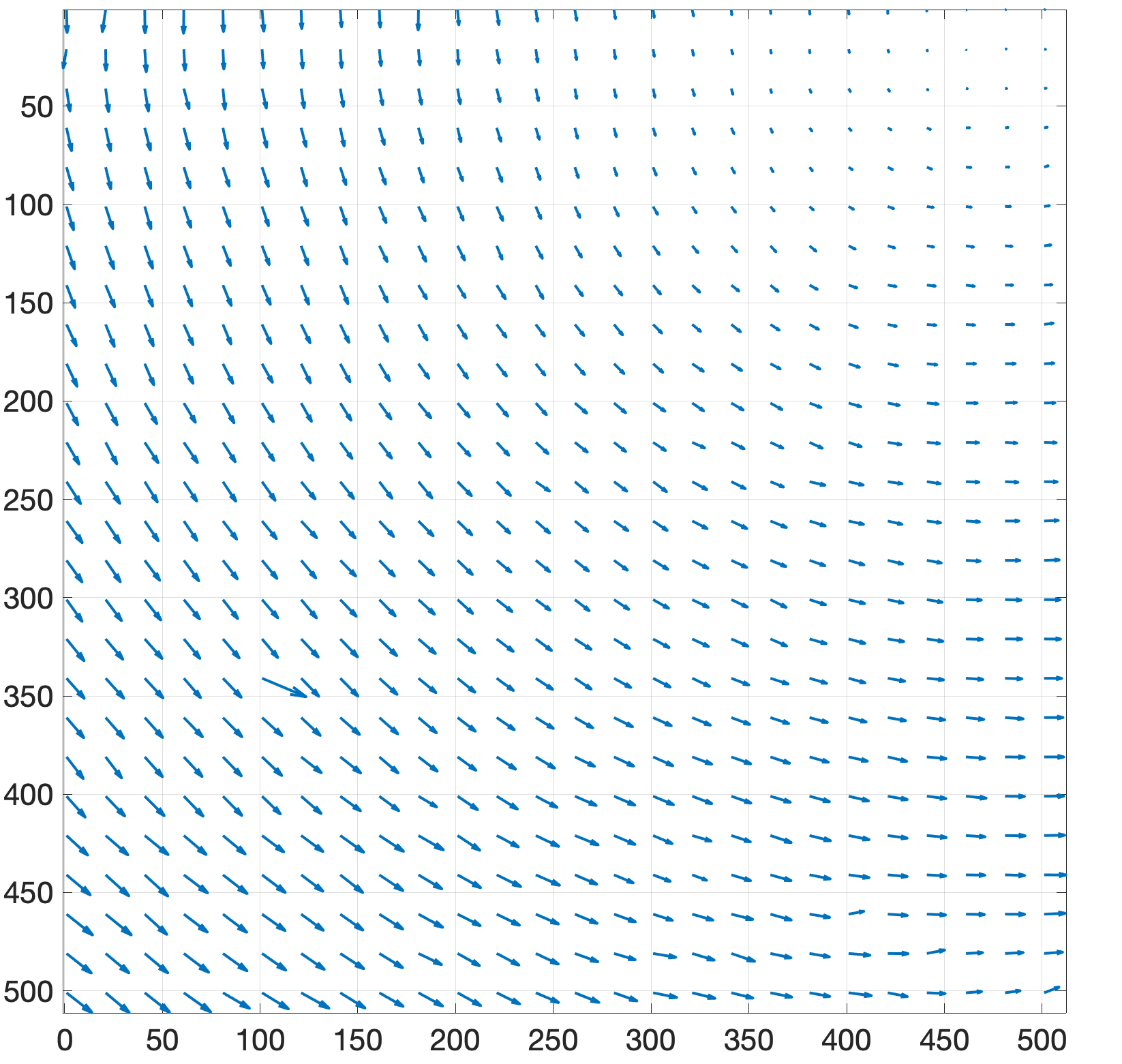} 
\caption{Quiver plot}\label{fig:Camera Motion Motivation 1 b}
\end{subfigure}
\begin{subfigure}[b]{.209\textwidth}
\includegraphics[width = .99\textwidth]{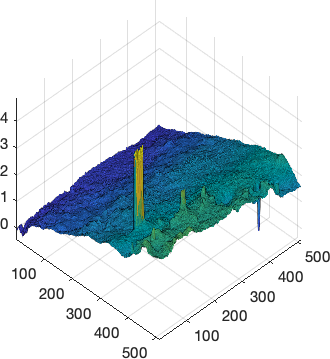} 
\caption{$x$ componenet}\label{fig:Camera Motion Motivation 1 c}
\end{subfigure}
\begin{subfigure}[b]{.28\textwidth}
\includegraphics[width = .99\textwidth]{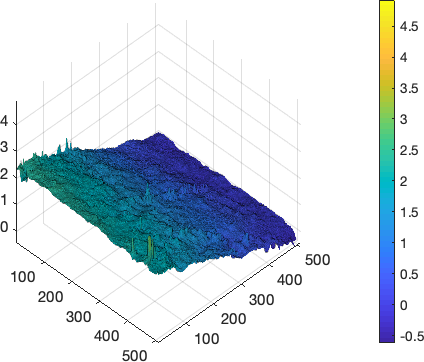} 
\caption{$y$ componenet}\label{fig:Camera Motion Motivation 1 d}
\end{subfigure}
\caption{Representations of the optical flow field corresponding to the \textit{Courtyard4} sequence at frame 30 without the inclusion of simulated turbulence.}\label{fig:Camera Motion Motivation 1}
\end{figure}

For the purposes of this work, global camera motion or simply, ``global motion'', will refer to the rigid body motion that a camera may undergo.
For generality, all angles of rotation and directions of translation are assumed to be possible.
In this section, we shall discuss the physical model that governs the impact of global camera motion on perceived optical flow in the absence of atmospheric turbulence, followed by several methods that to build a model to approximate the camera motion flow.
Finally, we provide a discussion on building a motion model when atmospheric turbulence is present.\\

The impact of camera motion in an optical flow field is apparent when looking at a color representation of a flow in Figure~\ref{fig:Camera Motion Motivation 1 a} and the vector field quiver plot in Figure~\ref{fig:Camera Motion Motivation 1 b}.
In order to investigate a model, we look to the $x$ and $y$ components of our optical flow field ${\bf V} = \left(V_x, V_y\right)$.
Each vector component will produce a 2D surface, hence we utilize a surface plot in order to visualize the results as shown Figure~\ref{fig:Camera Motion Motivation 1 c}-\ref{fig:Camera Motion Motivation 1 d}.
By investigating each component separately, we see that there are smooth underlying structures in the flow that we may later extract.
It is also important to notice that the moving object stands out as a peak in the $x$ component surface, thus any method that we choose must preserve the peak corresponding to the moving object.
Through the use of the pinhole camera model~\cite{trucco1998introductory}, we may derive the relation between real world coordinates $(\mathcal{X}, \mathcal{Y}, \mathcal{Z})$ and the pixel coordinates $(x,y)$ which upon differentiation in time yield a model for the optical flow field corresponding to camera motion.  This relation relation is determined to be
\begin{align}
V_x &= \frac{T_z x - T_x \foc}{\mathcal{Z}} + \omega_x\frac{xy}{\foc} - \omega_y\left( \foc + \frac{x^2}{\foc} \right) + \omega_z y \nonumber \\
V_y &= \frac{T_z y - T_y \foc}{\mathcal{Z}} + \omega_x\left( \foc + \frac{y^2}{\foc} \right) - \omega_y\frac{xy}{\foc} - \omega_z x
\label{eqn: Motion Equation}
\end{align}
where $\foc$ is the focal length, $T_{(x,y,z)}$ and $\omega_{(x,y,z)}$ correspond to translation and rotation movement in the $x,y,z$ direction.
This model is derived in detail in Appendix~\ref{sec: Motion Model}. Next, we present two methods which may extract the camera motion flow.
First, we take a physics based approach by exploiting the optical flow camera motion model in Equation~\eqref{eqn: Motion Equation}.
Second, we employ a filtering method to robustly obtain a camera motion flow model.

\subsection{Analytic Motion Model Derivation}\label{sec: Analytic method}
First, we propose a method of determining each parameter from the camera motion model by making an assumption of constant depth $\mathcal{Z}$.
Pseudo code for the algorithm proposed in this section is presented in Algorithm~\ref{alg: Analytic Background Algorithm}.
By taking the curl of Equation~\eqref{eqn: Motion Equation} and asserting that $V_z = 0$ since we consider the focal length to be fixed, we get
\begin{align}
\nabla \times {\bf V} =& {\bf \hat{i}}\left( -\partial_z V_y\right)  - {\bf \hat{j}}\left( -\partial_z V_x\right) + {\bf \hat{k}}\left( \partial_x V_y - \partial_y V_x\right) \nonumber \\
=&{\bf \hat{i}}\left( \frac{T_z y - T_y \foc}{\mathcal{Z}^2} \right) + {\bf \hat{j}}\left( \frac{ T_x \foc - T_z x}{\mathcal{Z}^2} \right) \nonumber \\
+&{\bf \hat{k}}\left( -\omega_y \frac{y}{\foc} - \omega_x\frac{x}{\foc} - 2\omega_z\right).
\end{align}
As the captured motion of interest is in the 2D image plane, we only need to consider the ${\bf \hat{k}}$ component, \textit{i.e.} 
\begin{equation}
(\nabla \times {\bf V})\cdot {\bf \hat{k}} = -\omega_y \frac{y}{\foc} - \omega_x\frac{x}{\foc} - 2\omega_z.
\end{equation}
As mentioned in Appendix~\ref{sec: Motion Model}, the image plane lies along the optical axis where its origin is at $(0, 0, \foc)$.  We note that the image domain is 
symmetric about the origin and ergo the average of the $x$ and $y$ coordinates is zero. Thus, denoting $\langle \cdot \rangle$ to be the spatial average value over the image 
domain we get,
\begin{equation}
\langle (\nabla \times {\bf V})\cdot {\bf \hat{k}} \rangle = \langle -\omega_y \frac{y}{\foc} - \omega_x\frac{x}{\foc} - 2\omega_z \rangle = -2\omega_z,
\end{equation}
which reveals
\begin{equation}
\omega_z = \frac{-1}{2}\langle (\nabla \times {\bf V})\cdot {\bf \hat{k}} \rangle.
\label{eqn: omega z}
\end{equation}
Next, taking the derivative with respect to each coordinate, we find
\begin{align}
\partial_x \left( (\nabla \times {\bf V})\cdot {\bf \hat{k}} \right)&= -\frac{\omega_x}{\foc} \nonumber \\
\partial_y \left( (\nabla \times {\bf V})\cdot {\bf \hat{k}} \right)&= -\frac{\omega_y}{\foc}.
\end{align}
After taking the spatial average we obtain
\begin{align}
\frac{\omega_x}{\foc} &= -\left\langle\partial_x \left( (\nabla \times {\bf V})\cdot {\bf \hat{k}} \right)\right\rangle \nonumber \\
\frac{\omega_y}{\foc} &= -\left\langle\partial_y \left( (\nabla \times {\bf V})\cdot {\bf \hat{k}} \right)\right\rangle.
\label{eqn: omega x,y}
\end{align}
This provides us all of the radial motion information.
Next, we discuss translation components $T_{x,y,z}$.
At this stage, we assume that the information from 
$\omega$ has already been computed.
By taking the spatial average of the velocity field,
\begin{align}
\langle V_x \rangle &= \langle V_x^T \rangle + \langle V_x^\omega \rangle = \langle V_x^T \rangle + \left\langle -\frac{\omega_y}{\foc}\left( \foc^2 + x^2\right) \right\rangle \nonumber\\
\langle V_y \rangle &= \langle V_y^T \rangle + \langle V_y^\omega \rangle = \langle V_y^T \rangle + \left\langle \frac{\omega_x}{\foc}\left( \foc^2 + y^2\right) \right\rangle .
\end{align}
where
\begin{align}
\langle V_x^T \rangle &= \left\langle \frac{T_z x - T_x \foc}{\mathcal{Z}} \right\rangle = \left\langle \frac{T_z x}{\mathcal{Z}}\right\rangle - \left\langle\frac{T_x \foc}{\mathcal{Z}} \right\rangle \nonumber \\
\langle V_x^T \rangle &= \left\langle \frac{T_z x - T_x \foc}{\mathcal{Z}} \right\rangle = \left\langle \frac{T_z y}{\mathcal{Z}}\right\rangle - \left\langle\frac{T_y \foc}{\mathcal{Z}} \right\rangle.
\end{align}
Assuming $\mathcal{Z}$ to be constant, and hence $\frac{T_x}{\mathcal{Z}}$, $\frac{T_y}{\mathcal{Z}}$ and $\frac{T_z}{\mathcal{Z}}$ is also considered as constant values given by,
\begin{align}
\langle V_x^T \rangle &= - \left\langle\frac{T_x \foc}{\mathcal{Z}} \right\rangle = - \frac{T_x \foc}{\mathcal{Z}}\nonumber \\
\langle V_y^T \rangle &= - \left\langle\frac{T_y \foc}{\mathcal{Z}} \right\rangle = - \frac{T_y \foc}{\mathcal{Z}},
\end{align}
or equivalently,
\begin{align}
\frac{T_x \foc}{\mathcal{Z}} &=  -\langle V_x^T \rangle =  \left\langle -\frac{\omega_y}{\foc}\left( \foc^2 + x^2\right) \right\rangle- \langle V_x \rangle \nonumber \\
\frac{T_y \foc}{\mathcal{Z}} &=  -\langle V_y^T \rangle =  \left\langle \frac{\omega_x}{\foc}\left( \foc^2 + y^2\right) \right\rangle - \langle V_y \rangle.
\label{eqn: T x,y}
\end{align}
Finally taking the divergence of our vector field we find the translation component $T_z$, \textit{i.e.} 
\begin{equation}
\nabla \cdot {\bf V} = 2\frac{T_z}{\mathcal{Z}} + 3\omega_x\frac{y}{\foc} - 3\omega_y\frac{x}{\foc}
\end{equation}
and in spatial average,
\begin{equation}
\left\langle \nabla \cdot {\bf V} \right\rangle = 2\left\langle \frac{T_z}{\mathcal{Z}} \right\rangle,
\end{equation}
hence, again considering that $\frac{T_z}{\mathcal{Z}}$ is constant, we obtain
\begin{equation}
\frac{T_z}{\mathcal{Z}} = \frac{1}{2} \left\langle \nabla \cdot {\bf V} \right\rangle.
\label{eqn: T z}
\end{equation}
These equations give us a closed form solution to find all parameters within the model with the assumption of a constant $\mathcal{Z}$.
From the numerical perspective, the derivatives are computed using a first order finite difference scheme.
Prior to computing the derivative, the data is smoothed in order to mitigate error induced by localized fluctuations.
The smoothing is performed with a 2D Gaussian filter on both $V_x$ and $V_y$ with a standard deviation equal to one seventh the number of columns and rows.
The value of one seventh is chosen as it removes localized spatial fluctuations, such as moving objects and atmospheric turbulence, and may be changed for different applications.
\begin{algorithm}
\caption{Pseudo code corresponding to the analytical compensation model}\label{alg: Analytic Background Algorithm}
\begin{algorithmic}[1]
\BState Inputs: flow to decompose ${\bf V}$, focal length $\foc$
\BState Smooth $V_x$ and $V_y$ with a Gaussian filter 
\BState Compute $\omega_z$ \eqref{eqn: omega z}, and $\omega_{x,y}$ \eqref{eqn: omega x,y}
\BState Using values $\frac{\omega_{x,y}}{\foc}$ compute $\frac{T_{x,y} \foc}{\mathcal{Z}}$ \eqref{eqn: T x,y}
\BState Compute $\frac{T_z}{\mathcal{Z}}$ \eqref{eqn: T z}
\BState Substitute the computed components into \eqref{eqn: Motion Equation} to produce the approximated global motion flow $\bf M$.
\BState Compute compensated optical flow ${\bf V}_{c}={\bf V}-{\bf M}$
\BState \textbf{return} ${\bf V}_{c}$
\end{algorithmic}
\end{algorithm}

\subsection{Empirical Camera Motion Model}\label{sec: Gaussian method}
In this section, we take an image processing standpoint to introduce a more robust empirical method to the camera motion flow as compared to the physically derived analytic method in the previous section.
We proceed with motivation from Equation~\eqref{eqn: Motion Equation}. Unfortunately, the assumption of a constant depth is not generally true, hence a more robust method is required.\\
Experimentally, when implementing a Gaussian filter as a preprocessing step for differentiation in the analytic method, the smoothed flow field produced appears qualitatively similar to the expected camera motion model that we aim to estimate.
This result is expected as the assumption of a smooth function $\mathcal{Z}(x,y)$, the model from \eqref{eqn: Motion model} will remain a smooth function even though it is no longer a simple polynomial.
Recalling the end goal of this work, detecting moving objects in atmospheric turbulence, we have no need for the parameters of motion and thus we take the desired motion model to correspond to the output of a low pass filter.
Taking a low pass filter will remove localized fluctuations in the optical flow resulting from atmospheric turbulence and object motion, thus leaving only the global camera motion.
Just as we did in the analytic method, we empirically take the standard deviation of the Gaussian to be one seventh the number of rows and columns in order to mitigate the impact of localized fluctuations.
The smoothing is performed separately on each component of the optical flow at each frame.
\begin{figure}[b]
	\centerline{
        \includegraphics[width = .25\linewidth]{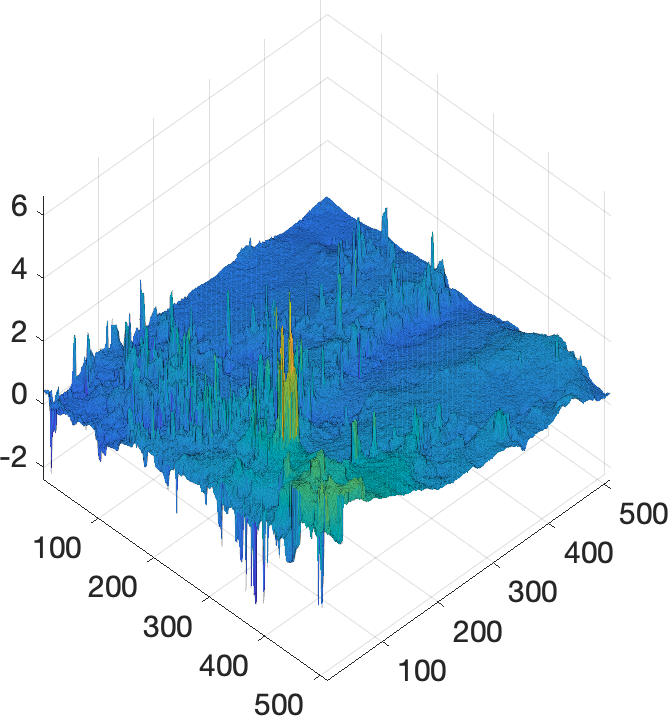}
		\hfill
        \includegraphics[width = .25\linewidth]{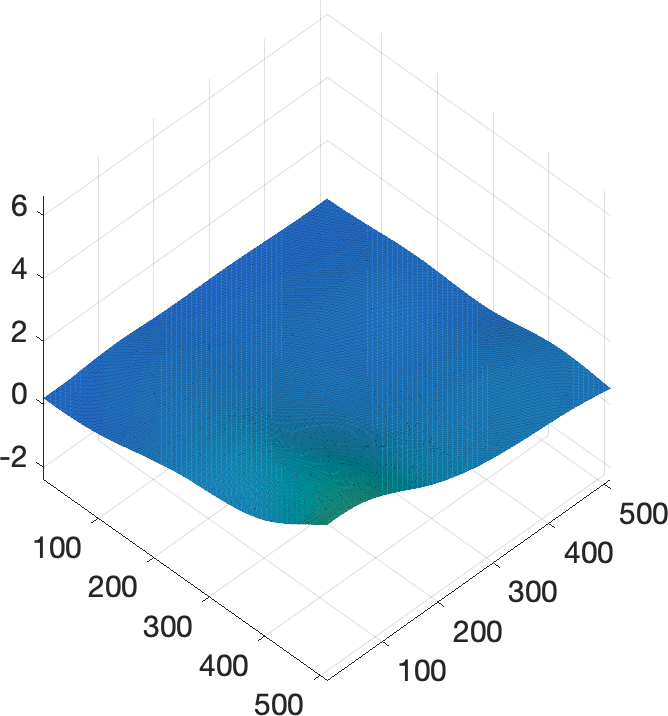}
        \hfill
        \includegraphics[width = .3\linewidth]{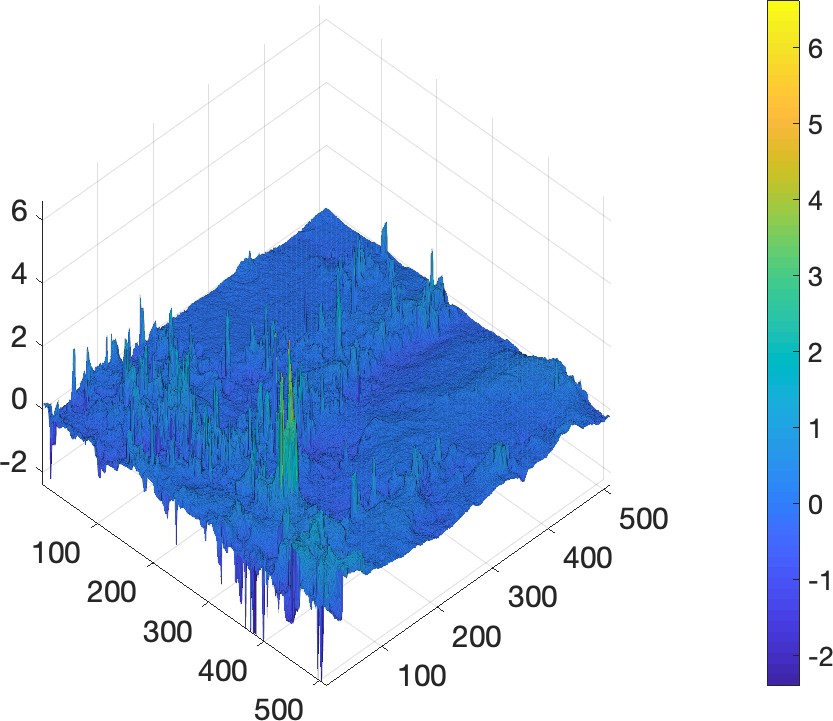}
	}
	\centerline{
		\parbox{.25\linewidth}{
		\centering
		Optical flow
		}
		\hfill
		\parbox{.25\linewidth}{
		\centering
		Motion model
		}
		\hfill
		\parbox{.3\linewidth}{
		\centering
		Compensated optical flow
		}
	}
	\centerline{
        \includegraphics[width = .25\linewidth]{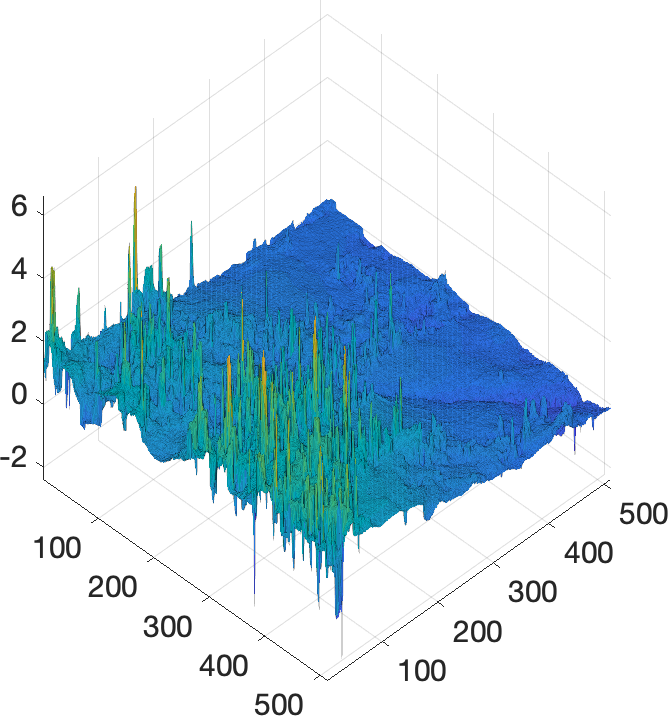}
		\hfill
        \includegraphics[width = .25\linewidth]{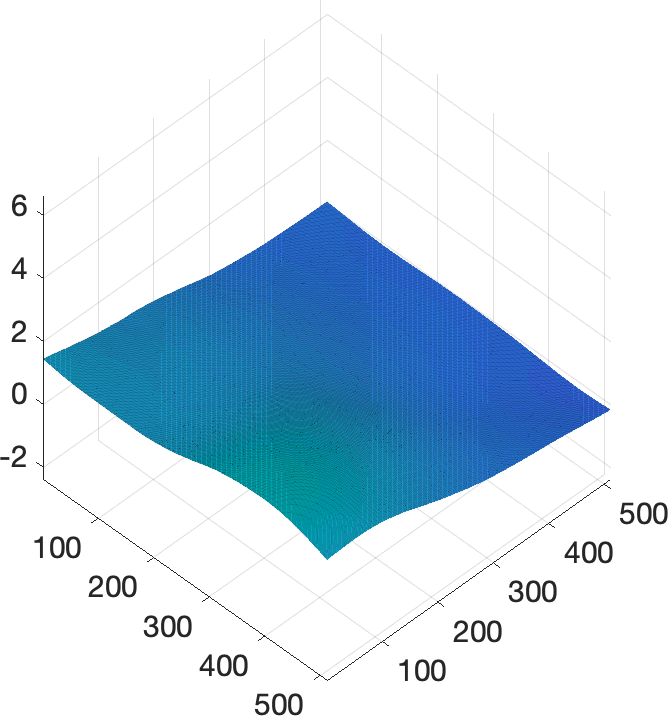}
        \hfill
        \includegraphics[width = .3\linewidth]{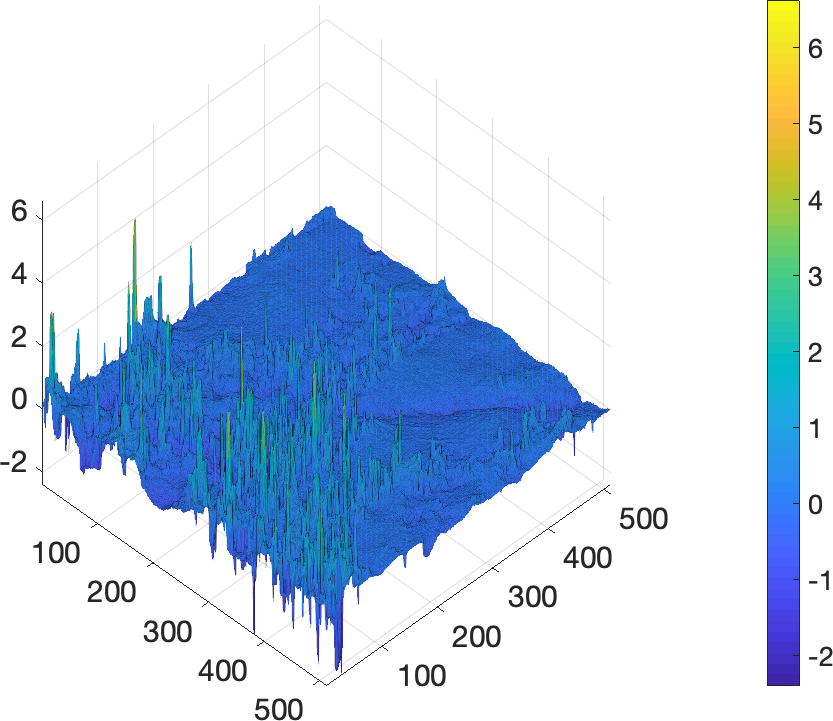}
	}
   \caption{Demonstration of empirical motion model results when applied to \textit{Courtyard4} at frame 30 with the inclusion of atmospheric turbulence. The $x$ and $y$ component are shown on the top and bottom left respectively. In the middle is the corresponding components of the camera motion model.  The compensated optical flow is shown for each component to the right.}
\label{fig:Surface Courtyard4 Flow Example}
\end{figure}

\subsection{In the Presence of Atmospheric Turbulence}\label{sec: Introduciton of Turbulence}
\begin{figure}[t]
	\centerline{
        \includegraphics[width = .23\linewidth]{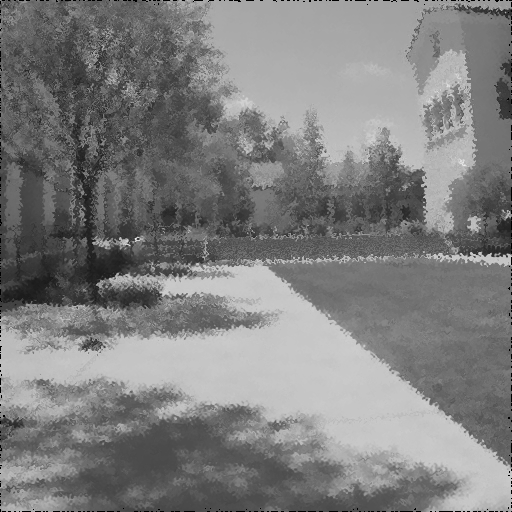}
		\hfill
        \includegraphics[width = .23\linewidth]{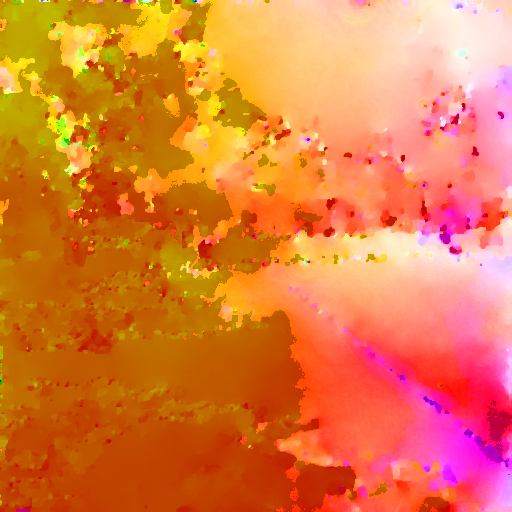}
        \hfill
        \includegraphics[width = .23\linewidth]{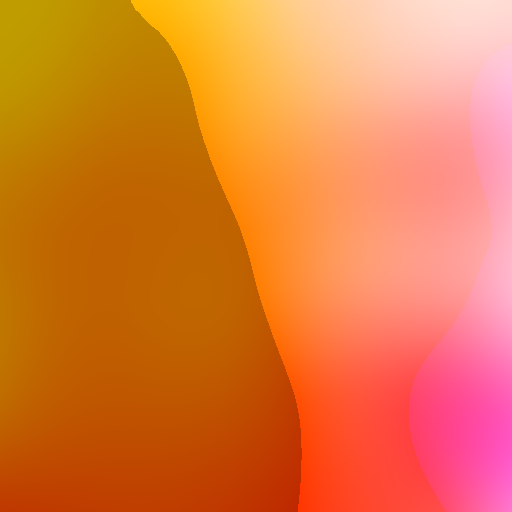}
        \hfill
        \includegraphics[width = .23\linewidth]{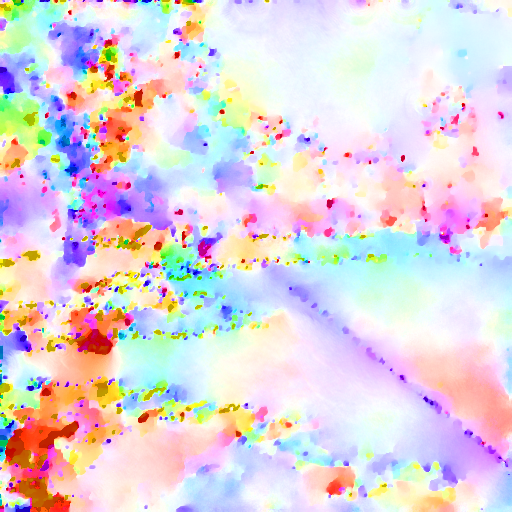}
		}
	\centerline{
	\parbox{.23\linewidth}{
	\centering
	Frame 30
	}
	\hfill
	\parbox{.23\linewidth}{
	\centering
	Optical flow $\bf V$
	}
	\hfill
	\parbox{.23\linewidth}{
	\centering
	Camera motion model $\bf M$
	}
	\hfill
	\parbox{.23\linewidth}{
	\centering
	Motion-compensated optical flow ${\bf V_c}$
	}
	\hfill
	}
\caption{Demonstration of camera motion subtraction method from \textit{Courtyard4} at frame 30 with simulated atmospheric turbulence.  On the left, we see the image corresponding to frame 30 which has observable deformation due to simulated atmospheric turbulence. To its right, we see the corresponding optical flow where the impact of atmospheric turbulence is seen with the majority of the flow field occluded by camera motion. Next, we have the camera motion model that is computed by the empirical method. Finally, we have the camera motion-compensated optical flow field on the right, where we see only the impact of atmospheric turbulence.}
\label{fig:Compensated Flow Example}
\end{figure}

Since the main goal of this work is to detect objects in a flow field impacted by both atmospheric turbulence and global camera motion, we now need to look to the case where we include atmospheric turbulence.
We will simulate atmospheric turbulence on the example from \textit{Courtyard4} in Figure~\ref{fig:Camera Motion Motivation 1} so we may compare our result to the original sequence without atmospheric turbulence.
In practice, the analytic method from Section~\ref{sec: Analytic method} is unable to consistently account for camera motion observed in the sequence when turbulence is added as we loose the ability to take reliable derivatives due to the random motion induced from atmospheric turbulence.
The empirical filtering method from Section~\ref{sec: Gaussian method} is able to handle the localized oscillations from the atmospheric turbulence by removing them through smoothing and hence is used routinely in this work.
An example of the motion model is demonstrated in Figure~\ref{fig:Surface Courtyard4 Flow Example} where we see that we indeed remove the motion that is introduced from camera motion as the resulting compensated optical flow is centered about zero.
A colorized depiction of the flow field is shown in Figure~\ref{fig:Compensated Flow Example} where we see the orange and red motion from the camera motion is removed leaving a primarily white background.

\section{Flow Field Decomposition}\label{sec: Cartoon+Texture Decomposition}
\begin{figure}[b]
\centering
   	\includegraphics[width = .49\linewidth]{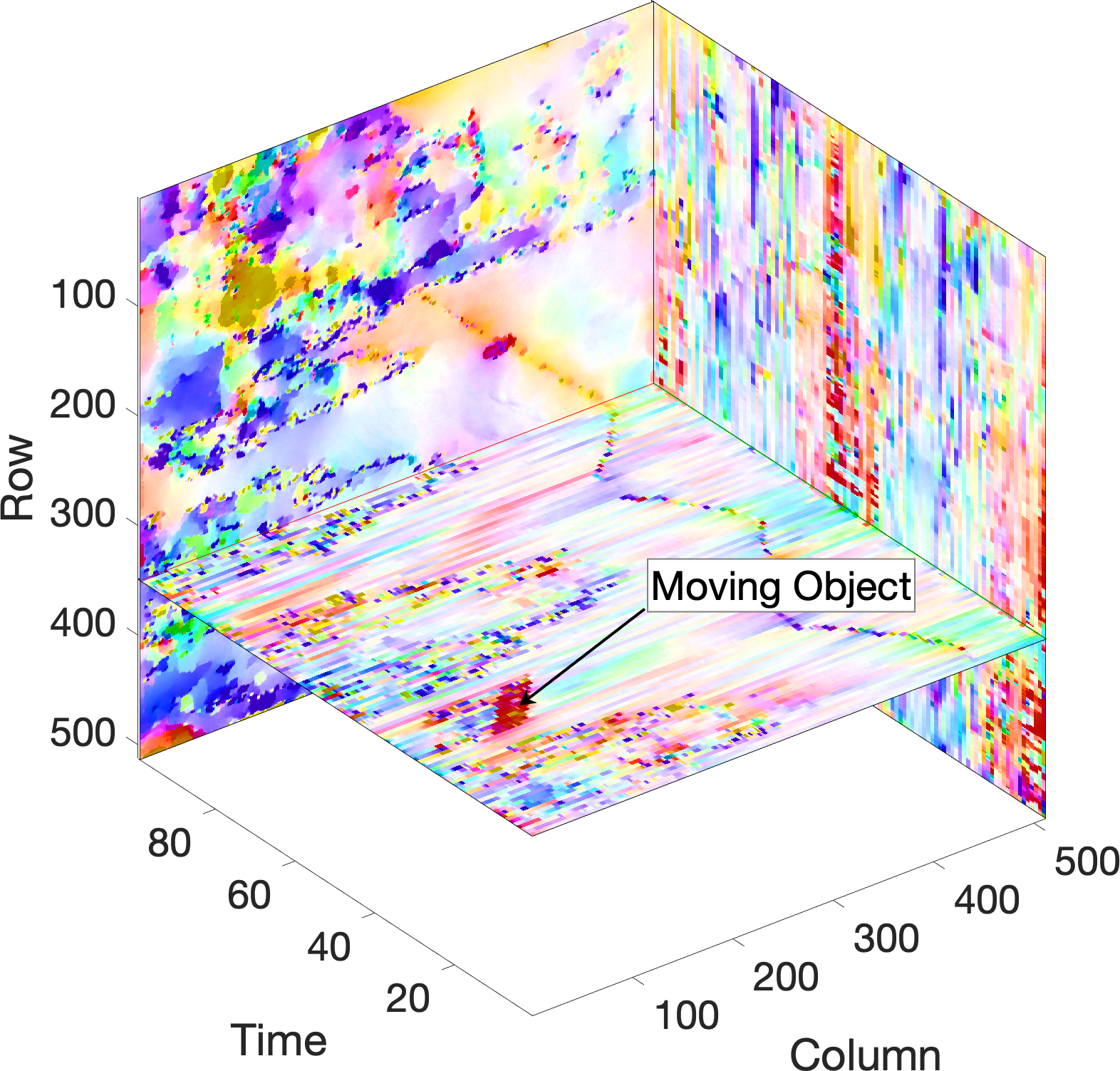}
   	\caption{Visualization of the temporal homogeneity that arises from a moving object in the compensated optical flow from \textit{Courtyard4} using the Slicer application.}\label{fig:Compensated Flow Slicer}
\end{figure}

In this section, we address the problem of decomposing a 2D spatio-temporal vector field into its non-oscillating and oscillating components.
We will proceed by using the empirical method described in Section~\ref{sec: Gaussian method} to remove the camera motion flow field as it produces the most consistent and reliable results.
Moving forward, we refer to Figure~\ref{fig:Compensated Flow Slicer} as our motivation for the proposed method.
In this figure we see that, even in the presence of atmospheric turbulence, a moving object will correspond to a region of locally homogeneous vectors when we take a 2D+Time viewpoint on the problem.
When viewed frame by frame, these homogeneous regions may not reveal a moving object, but using temporal information, these regions become apparent.

\subsection{Theoretical Background}
A solution of this problem was proposed in our previous work \cite{gilles2018detection} by using a decomposition model inspired by the classic ``cartoon+textures'' model used in image processing.
The cartoon part corresponds to the geometric information while the texture part corresponds to the oscillating component. 
This decomposition was extended to 2D spatio-temporal vector fields, \textit{i.e.} our camera motion compensated optical flow ${\bf V}_c=\left(V_{c_x}(x,y,t),V_{c_y}(x,y,t)\right)$. The idea is first to rewrite the vector field as a complex scalar field, \textit{i.e.} $\tilde{V}(x,y,t)=V_{c_x}(x,y,t)+\imath V_{c_y}(x,y,t)$.
Then we use both complex wavelet or curvelet transform~\cite{Candes2005} based on the computational or burden or resolution needs. The curvelet vector field decomposition model is given by
\begin{equation}
(\hat{u}, \hat{v}) = \argmin_{\substack{u\in \dot{C}_{1, 1}^{1}\\ v\in \dot{C}_{-1, \infty}^{\infty}}} \|u\|_{\dot{C}_{1, 1}^{1}} +J^{\ast}_{C}\left( \frac{v}{\mu}\right) + 
\frac{1}{2\lambda}\|I - (u+v)\|_{L^2}^2,
\label{eqn: Curvelet Model}
\end{equation}
where $\dot{C}_{1,1}^1$ is a curvelet based function space and $\dot{C}_{-1,\infty}^\infty$ its dual.
This model is utilized in order to preserve geometric information whereas changing to a wavelet based method will provide appreciably faster run time, as demonstrated in Figure~\ref{fig: CurveletWavelet Runtime}, at the cost geometric resolution.
The corresponding vector fields associated with $\hat{u}$ and $\hat{v}$ are respectively obtained by ${\bf u}=(\Re(\hat{u}),\Im(\hat{u}))$ and ${\bf v}=(\Re(\hat{v}),\Im(\hat{v}))$ where $\Re$ and $\Im$ denote real and imaginary components. If we denote
$$
\forall z = |z|e^{\imath \theta}\in \mathbb{C}, \quad CShrink(z, \lambda) = \max(0, |z|-\lambda)e^{\imath \theta},
$$
and $\mathcal{C}$ the curvelet transform,
then the decomposition pseudocode is given by Algorithm~\ref{alg: Wavelet Thresholding}.

\begin{algorithm}[h!]
\caption{Besov based decomposition numerical algorithm}
\label{alg: Wavelet Thresholding}
\begin{algorithmic}[1]
\BState Inputs: flow to decompose ${\bf V_c}$, parameters $\lambda, \mu$, maximum number of iterations $N_{max}$
\BState Initialization $n = 0,\tilde{V}=V_{c_x}+\imath V_{c_y},u^0 = 0,v^{0} = 0$
\BState \textbf{repeat} 
\State	$v^{n+1} = \tilde{V} - u^n - \mathcal{C}^{-1}(CShrink(\mathcal{C}(\tilde{V}-u^{n}), 2\mu))$
\State	$u^{n+1} = \mathcal{C}^{-1}(CShrink(\mathcal{C}(\tilde{V}-v^{n}), 2\lambda))$
\BState \textbf{until} $\max(\|u^{n+1} - u^{n}\|_{L^2}, \|v^{n+1} - v^{n}\|_{L^2})<10^{-4}$ or $n=N_{max}$
\BState \textbf{return} $\left(\Re(u^{n+1}),\Im(u^{n+1})\right), \,\left(\Re(v^{n+1}),\Im(v^{n+1})\right)$
\end{algorithmic}
\end{algorithm}

\subsection{Experimental Results}\label{sec: Experimental results}
\begin{figure}[t]
\centering
\hfill
\includegraphics[width=.49\linewidth]{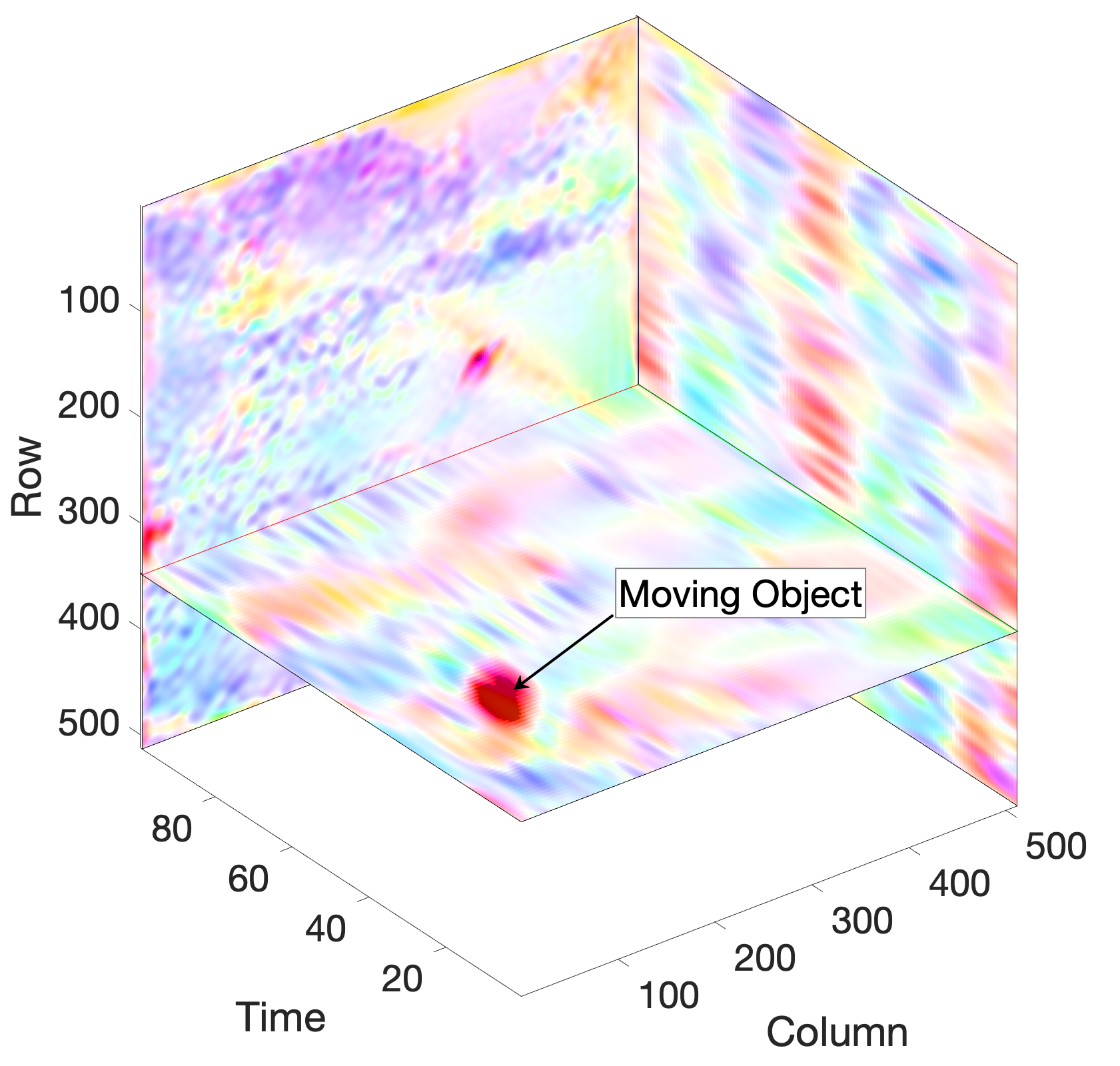} 
\hfill
\includegraphics[width=.49\linewidth]{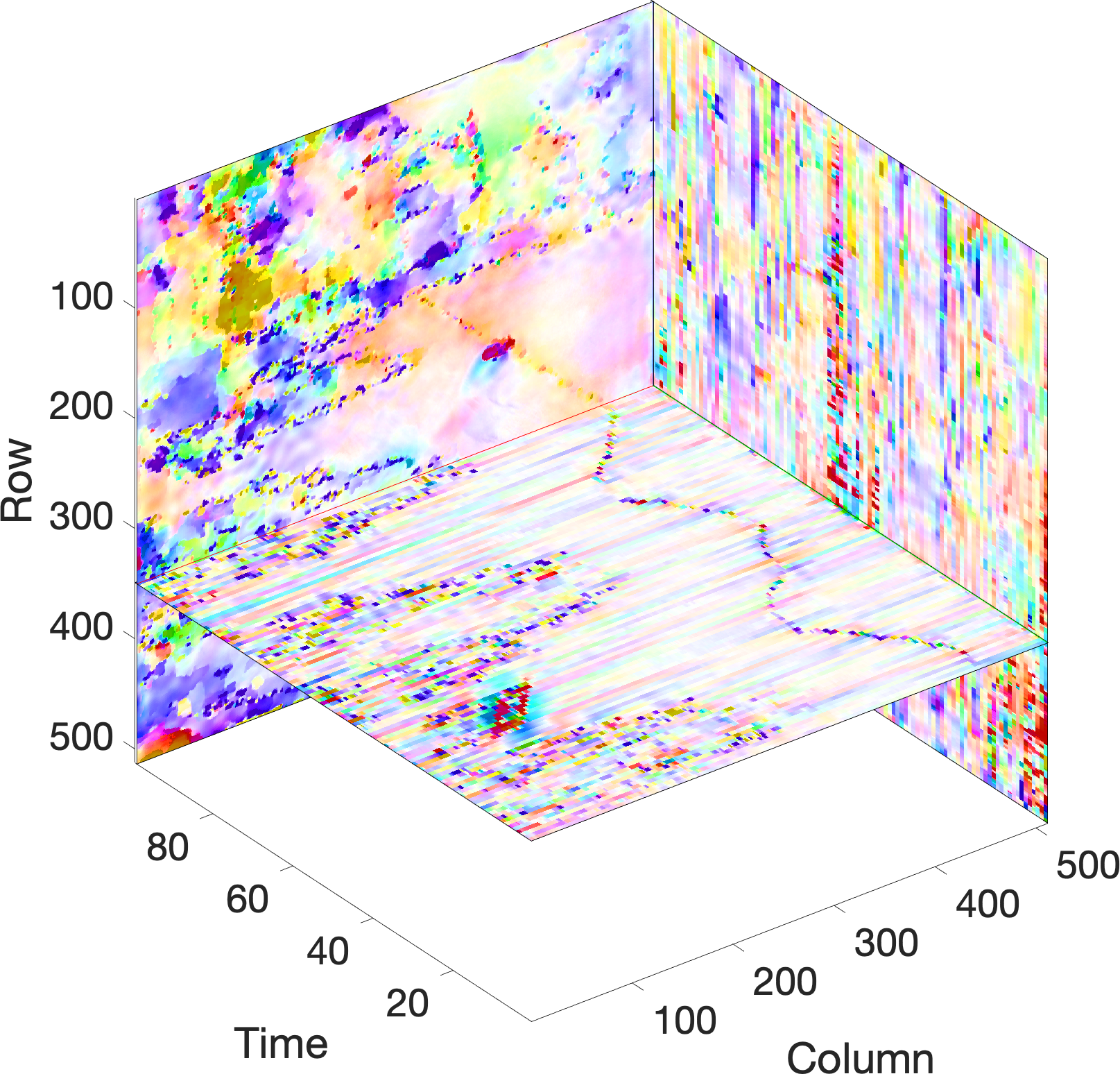} 
\hfill
\caption{Temporal view for the decomposition of the motion compensated optical flow from Figure~\ref{fig:Compensated Flow Slicer}. On the left is the geometric component of the optical flow and on the right is the oscillatory component. It is visually apparent that the moving object stands out in the geometric component, while the atmospheric turbulence is left in the oscillatory component.}
\label{fig: Courtyard 4 Curvelet Slicer}
\end{figure}
\begin{figure}[t]
\centering
\includegraphics[width=.7\linewidth]{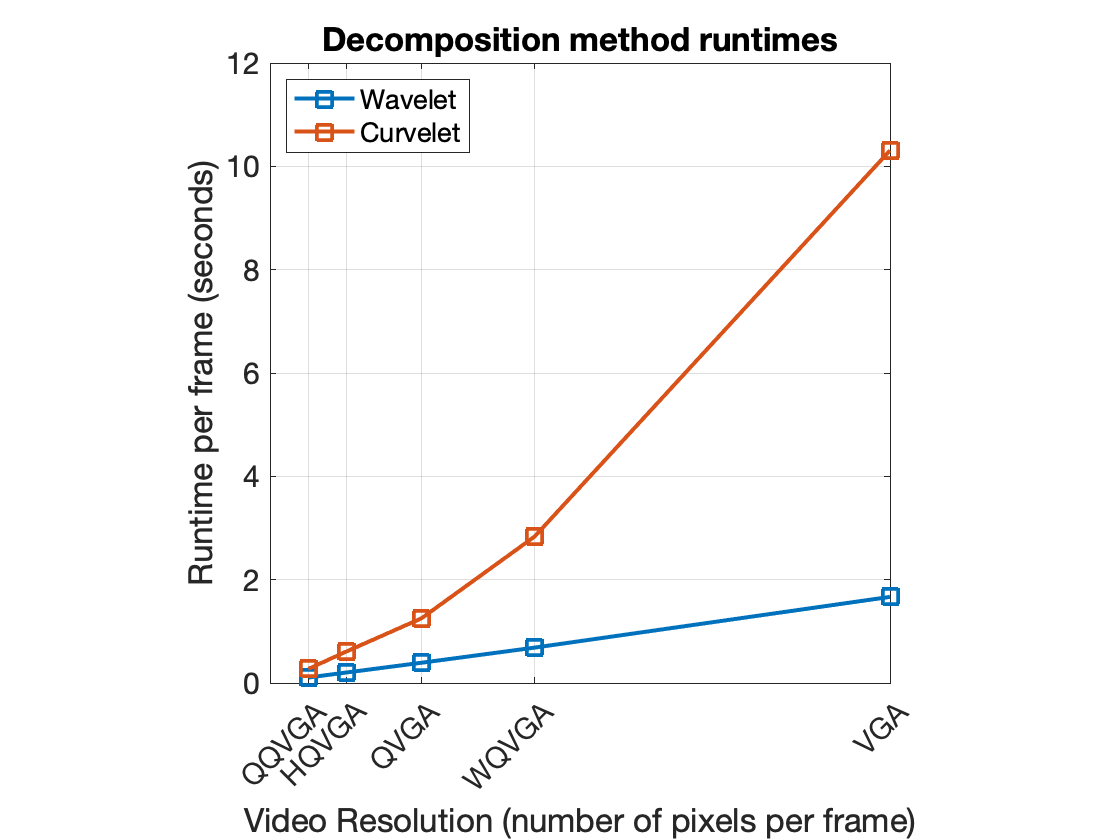} 
\caption{Comparison of a per-frame run time for the wavelet and curvelet decomposition for varying standard video sizes.}
\label{fig: CurveletWavelet Runtime}
\end{figure}

The decomposition algorithm was implemented in MATLAB and 
either uses the wavelet transform or the curvelet transform provided in the freely available Curvlab toolbox\footnote{\url{http://www.curvelet.org/software.html}}.  
Experimentally, the choice of $\mu=\lambda=1$ in the decomposition models works well for all sequences and is consistent with the results reported in \cite{gilles2018detection}. 
The maximum number of iterations 
was fixed to five, though fewer iterations are observed to attain the threshold $\max(\|u^{n+1} - u^{n}\|_{L^2}, \|v^{n+1} - v^{n}\|_{L^2})<10^{-4}$.
\begin{algorithm}[b]
\caption{Leave-one-out cross validation for outliers in optical flow}\label{alg: leave one out}
\begin{algorithmic}[1]
\BState Inputs: flow ${\bf F}$, integer threshold $\kappa$
\BState Initialization: Compute magnitude of the optical flow $\|{\bf F}_k\|$ at each frame $k=1,2,...,N_{\text{frames}}$
\BState \textbf{for }$i=1:N_\text{frames}$
\State Compute mean $M_i = \frac{1}{N_{\text{frames}}-1}\sum_{j\neq i}\|{\bf F}_{j}\|$
\State \textbf{end} 
\State compute overall mean $\tilde{M}$ and standard deviation $\tilde{S}$ for $\tilde{M}$,
$$
\tilde{M}=\frac{1}{N_{\text{frames}}}\sum_{l=1}^{N_{\text{frames}}}M_l, \quad \tilde{S}=\sqrt{\frac{1}{N_{\text{frames}}-1}\sum_{m=1}^{N_{\text{frames}}}|\tilde{M}-M_m|}
$$
\State inds $\leftarrow$ $\left\lbrace i :  \left|M_i - \tilde{M}\right| > \kappa \tilde{S}\right\rbrace$
\BState \textbf{return} inds
\end{algorithmic}
\end{algorithm}
Before we decompose the flow field, we utilize a leave-one-out cross validation~\cite{kohavi1995study} on the magnitude of the motion compensated optical flow to determine frames containing outliers in the optical flow computation.
As outlined in Algorithm~\ref{alg: leave one out}, this method is performed by first computing the maximum magnitude of the flow at each frame in the sequence.
Then, each maximum is removed from the data set and the change in the mean is observed.
Any change that is outside of five-standard deviations of mean is considered as an outlier in the computation in the optical flow.
Once the cross validation is performed any frames determined to be containing errors are linearly interpolated in time with neighboring frames deemed free of outliers.
As errors in optical flow computation appear in regions with significantly larger magnitude, these regions act as Dirac-delta functions in the flow field.
We know that the delta function is the neutral element in convolution and thus when using the wavelet decomposition the wavelet profile appears at the position of the outlier.
This property leaves detection throughout the flow field impossible as the wavelet corresponding to the error is large enough in magnitude to be considered a valid moving object.
The leave-one-out cross validation is performed with the function with $\kappa=5$ as it produces consistent satisfactory results.
Figure~\ref{fig: Courtyard 4 Curvelet Slicer} illustrates the decomposition of the motion compensated optical flow from Figure~\ref{fig:Compensated Flow Slicer} into its geometric and oscillatory components with the curvelet algorithm.
Using this perspective, we qualitatively demonstrate the impact of our method.
A quantitative analysis is left to Section~\ref{chap: Detection and tracking} where we implement a detection and tracking algorithm.
When the curvelet algorithm is implemented, it returns excellent results, but at a great computational cost.
Taking note that Algorithm~\ref{alg: Wavelet Thresholding} is valid for both wavelet and curvelet transforms, we implement both methods in this work.
We observe that implementing the wavelet transform gives similar qualitative results in the colorized optical flow as the curvelet transform, but a considerable drop in runtime is reached.
In Figure~\ref{fig: CurveletWavelet Runtime}, the speedup of the wavelet over the curvelet transform is emphasized with varying video resolutions.
A quantitative comparison of the wavelet and curvelet tranform detection and tracking results are shown in Section~\ref{chap: Detection and tracking} with quantitative metric scores.
In general application, the use of wavelet or curvelet decomposition provide similar results, but if one wishes to maintain the most geometric information, the curvelet decomposition will provide the best results.

\section{Detection \& Tracking}\label{chap: Detection and tracking}
After the flow field decomposition is performed, the next procedure is the detection and tracking of the moving object.
In order to detect a moving object in a flow field, we implement a velocity thresholding scheme as explained in Section~\ref{sec: Object Detection}.
Once the detection step is completed, the tracking step is implemented via a Kalman Filter \cite{kalman1960new}.

\subsection{Object Detection}\label{sec: Object Detection}
\begin{figure}[t]
\centerline{
\hfill
\includegraphics[width=.35\textwidth]{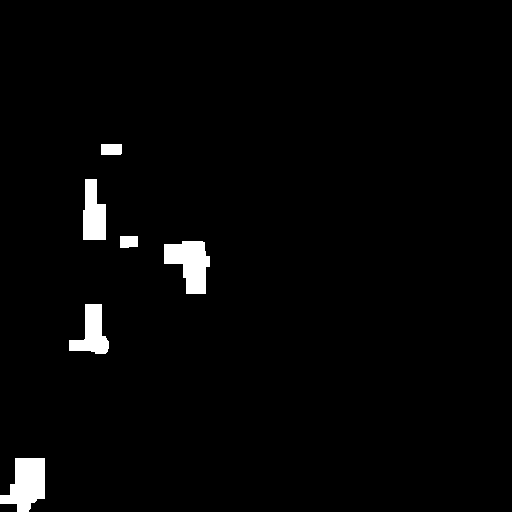}
\hfill
\includegraphics[width=.35\textwidth]{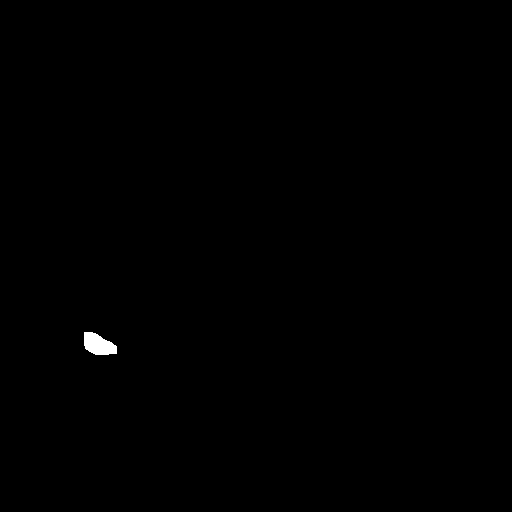}
\hfill
}
\centerline{
\hfill
\parbox{.45\textwidth}{\centering Motion Compensated Optical Flow}
\hfill
\parbox{.45\textwidth}{\centering Geometric Component}
\hfill
}
\centerline{
\hfill
\includegraphics[width=.45\textwidth]{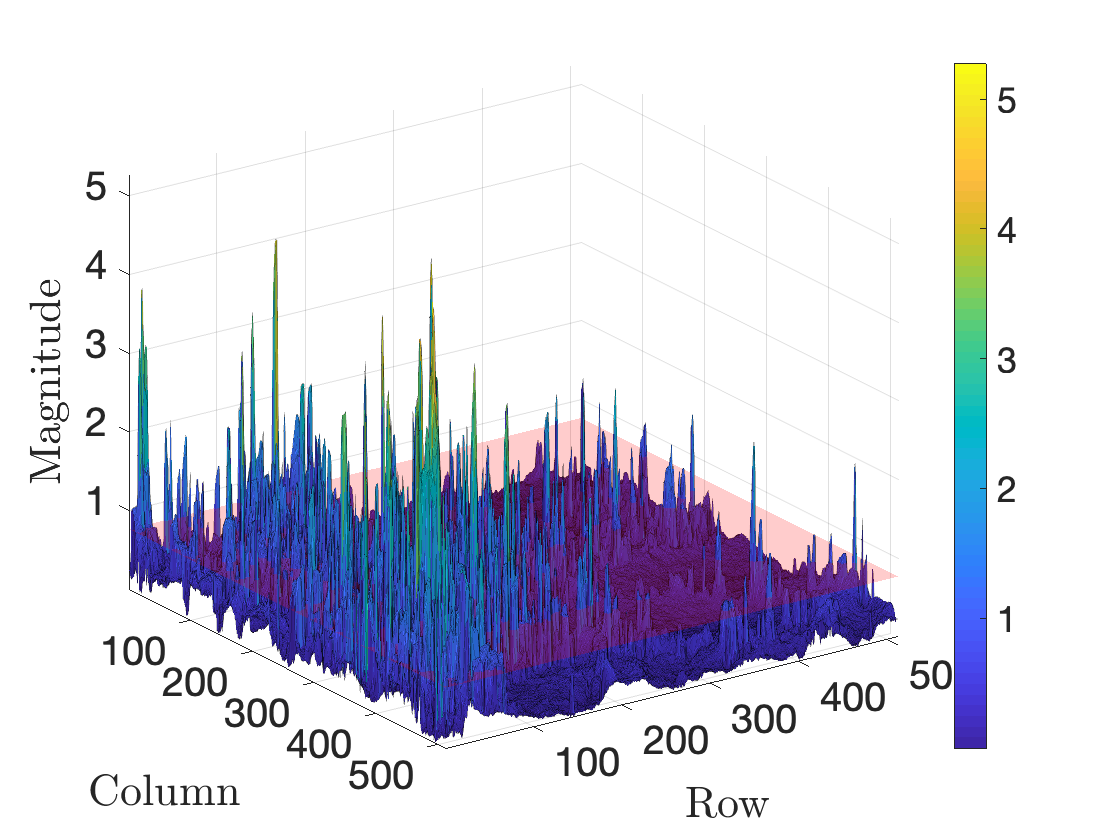}
\hfill
\includegraphics[width=.45\textwidth]{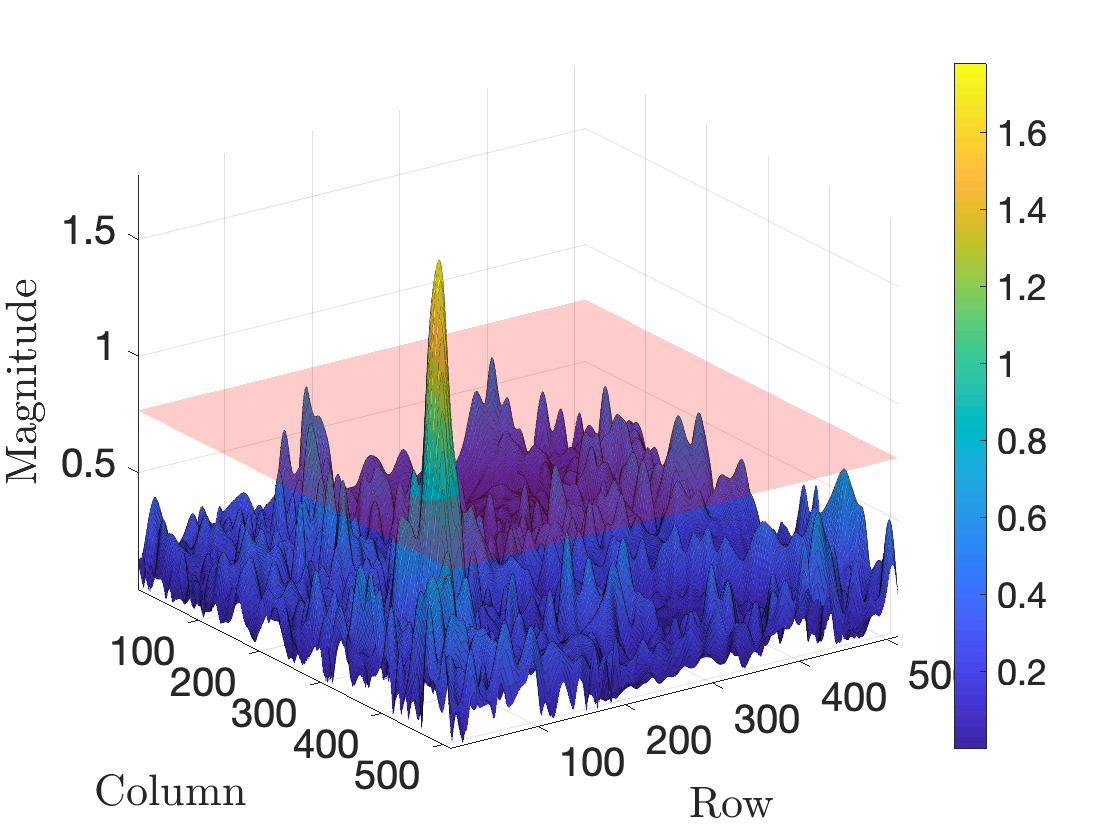}
\hfill
}
\caption{Detection and tracking result for \textit{Courtyard4} at frame 30 with the top presenting the detected mask and the bottom is the magnitude of the magnitude of the flow with the computed threshold shown in red.}
\label{fig: Courtyard 4 detection results}
\end{figure}

In order to detect moving objects in the geometric component of our flow field decomposition, we need to recall that parts of the low frequency oscillations from the atmospheric turbulence will remain in the geometric component.
With this thought in mind, the remaining low frequency oscillations captured in the geometric component will not correspond to large values (typically close to zero), but will be detected if a naive threshold operation is performed.
We see from a surface plot of the data and a color representation of the flow in Figure~\ref{fig: Courtyard 4 detection results}, that the moving object stands out from its surroundings in the geometric component of the flow than in the motion compensated optical flow.
In order to determine a threshold value, we compute a mean and a standard deviation of a set number of past frames (typically set to five), then set the threshold to be five standard deviations away from the mean value.
Implementing this procedure, we have the threshold value $T_i$ at each frame $i$, to be $T_i=\mathfrak{M}_i+5\sigma_{i}$ where $\mathfrak{M}_i$ is the mean magnitude and $\sigma_i$ the standard deviation of the magnitude of a flow $\bf F$.
In practice we compute $\mathfrak{M}_i$ and $\sigma_i$ through
$$
\mathfrak{M}_i=\frac{1}{N_{\text{rows}}N_{\text{cols}}}\sum_{r=1}^{N_{\text{rows}}}\sum_{c=1}^{N_{\text{cols}}} \| {\bf F}(r,c,i) \|,
\quad 
\sigma_i=\sqrt{\frac{1}{N_{\text{rows}}N_{\text{cols}}-1}\sum_{r=1}^{N_{\text{rows}}}\sum_{c=1}^{N_{\text{cols}}} |\| {\bf F}(r,c,i) \| - \mathfrak{M}_i|}.
$$
where $N_{\text{rows}}$ and $N_{\text{cols}}$ are the number of rows and columns in a frame, respectively.
Once the detection method is implemented, mathematical morphology operations are applied on the mask through image opening and closing.
Finally, the centroid and bounding box of each region are computed in the mask.

\subsection{Object Tracking}
With centroids in hand, the detected locations are assigned to existing tracks through the  Hungarian assignment Algorithm~\cite{munkres1957algorithms} using the optimized algorithm from Miller, Stone and Cox~\cite{miller1997optimizing}.  Once the detections are assigned to tracks, a prediction is made for each track by way of a Kalman Filter \cite{kalman1960new}.
In order to assign detections to tracks, the Kalman Filter requires one of two motion models: constant velocity or constant acceleration.
If the motion is assumed to be linear then the constant velocity is the best choice, and if the motion is assumed nonlinear then the constant acceleration is the better option.
We assume that in this application the motion fr
om the camera will be coupled with that of a moving object, hence a nonlinear motion model is a preferred choice.
Proceeding, any point that has passed the acceptance criterion is assigned to a track, and is annotated on both the sequence frame and its mask.
The tracking algorithm then proceeds to the next frame and continues until termination.
The results from applying the proposed method are shown in Figures~\ref{fig: Detection Courtyard 2}-\ref{fig: Detection Field1} where sequences with both simulated atmospheric turbulence and naturally forming atmospheric turbulence are demonstrated.

Each sequence is scored with a performance metric for the curvelet (${\bf u}_{\mathcal{C}}$), wavelet (${\bf u}_{\mathcal{W}}$), and compensated optical flow ${V}_{c}$ (\textit{i.e.} without any decomposition) and are shown in a table below each figure.
In each table, five different metrics are computed in order to provide a quantitative analysis.
Next to each metric we put an $\uparrow$ to denote the best score possible being one and $\downarrow$ being zero.
In this work we utilize the following metrics: the F-1 Score (F1$\uparrow$), positive predictive value (PPV$\uparrow$), false discovery rate rate (FDR$\downarrow$), accuracy (ACC$\uparrow$) and  false negative rate (FNR$\downarrow$).
Detailed information on the computation of these metrics is presented by Fawcett~\cite{fawcett2006introduction}.
Due to the amount of generated atmospheric turbulence, the camera motion compensated flow field is unable to discern the moving object from its surroundings.
However, once the geometric component of the same field is investigated, the moving object is located and tracked.
In Figure~\ref{fig: Detection Field1}, due to the lack of atmospheric turbulence that was formed, both algorithms perform comparably. The performance analysis is computed by providing a ground truth structure from the ground truth application mentioned in Section~\ref{sec: Data Collection}.

\newlength{\figSize}
\setlength{\figSize}{.11\textwidth}
\begin{figure}[p!]
\centering
\fbox{\begin{minipage}{.95\linewidth}
\centerline{
\hfill
\parbox{\figSize}{\centering Frame 20}
\hfill
\parbox{\figSize}{\centering Frame 35}
\hfill
\parbox{\figSize}{\centering Frame 50}
\hfill
\parbox{\figSize}{\centering Frame 65}
\hfill
\parbox{\figSize}{\centering Frame 80}
\hfill
\parbox{\figSize}{\centering Frame 95}
\parbox{.06\textwidth}{\centering ~}
}   
\end{minipage}}
\fbox{\begin{minipage}{.95\linewidth}
	\centerline{
		\hfill
        \includegraphics[width = \figSize]{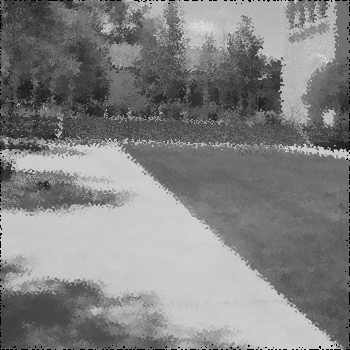}
		\hfill
        \includegraphics[width = \figSize]{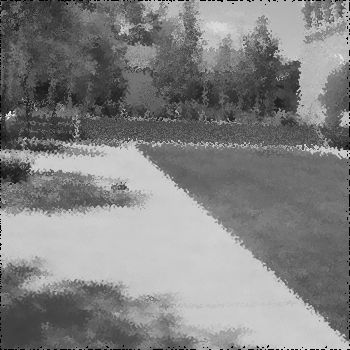}
		\hfill
		\includegraphics[width = \figSize]{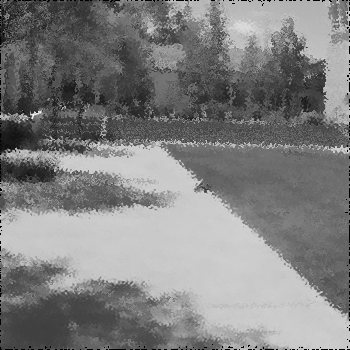}
		\hfill
		\includegraphics[width = \figSize]{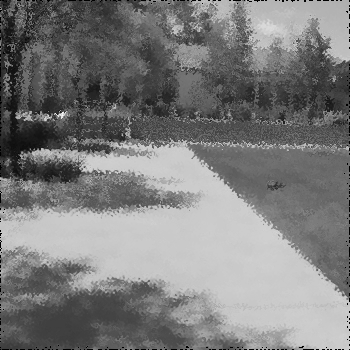}
		\hfill
		\includegraphics[width = \figSize]{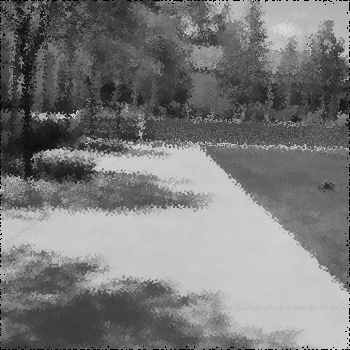}
		\hfill
		\includegraphics[width = \figSize]{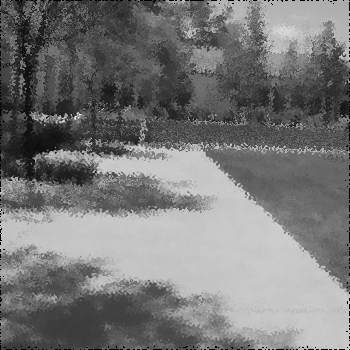}
		\parbox{.06\textwidth}{\centering ~}
		}
	\centerline{
		\hfill
        \includegraphics[width = \figSize]{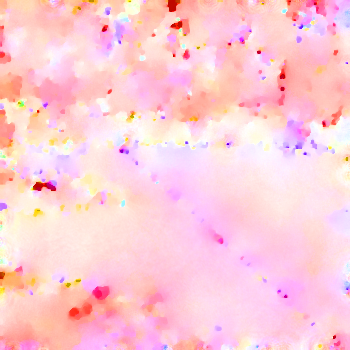}
		\hfill
        \includegraphics[width = \figSize]{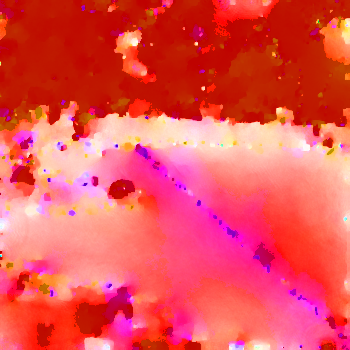}
		\hfill
		\includegraphics[width = \figSize]{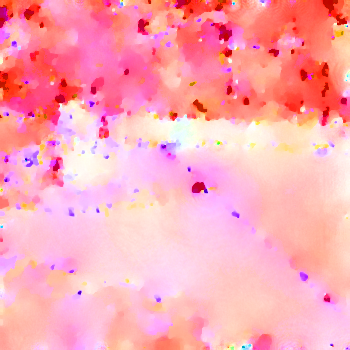}
		\hfill
		\includegraphics[width = \figSize]{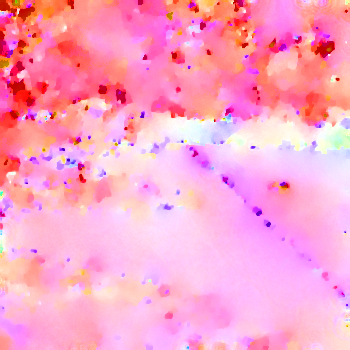}
		\hfill
		\includegraphics[width = \figSize]{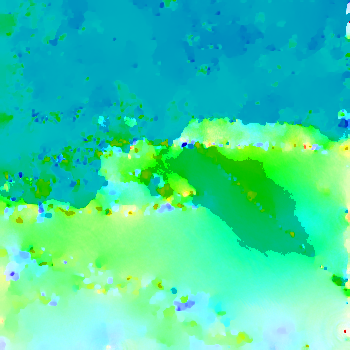}
		\hfill
		\includegraphics[width = \figSize]{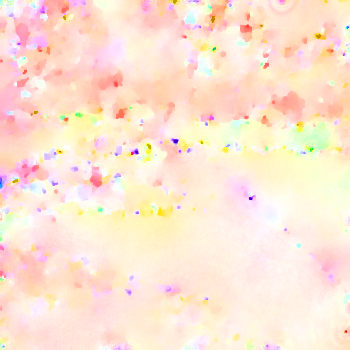}
		\parbox{.06\textwidth}{\centering \fbox{${\bf V}$}}
		}
\end{minipage}}
\fbox{\begin{minipage}{.95\linewidth}
	\centerline{
		\hfill
        \includegraphics[width = \figSize]{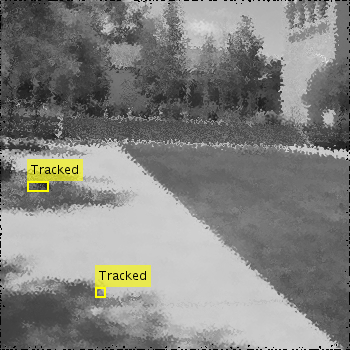}
		\hfill
        \includegraphics[width = \figSize]{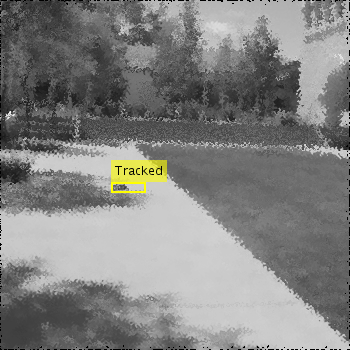}
		\hfill
		\includegraphics[width = \figSize]{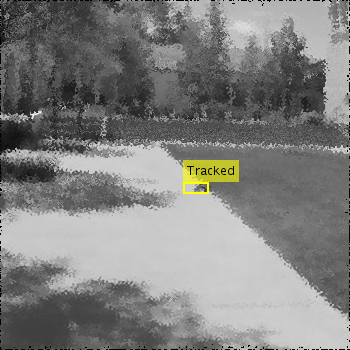}
		\hfill
		\includegraphics[width = \figSize]{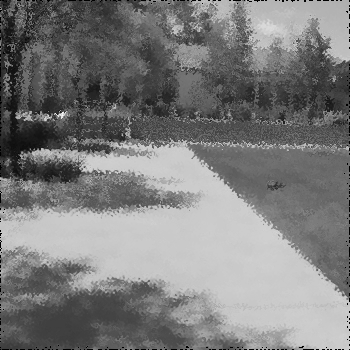}
		\hfill
		\includegraphics[width = \figSize]{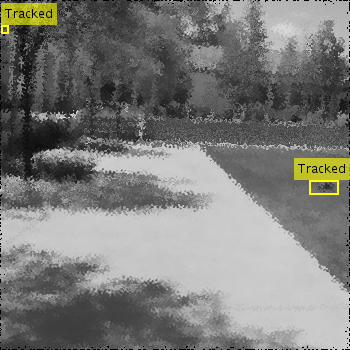}
		\hfill
		\includegraphics[width = \figSize]{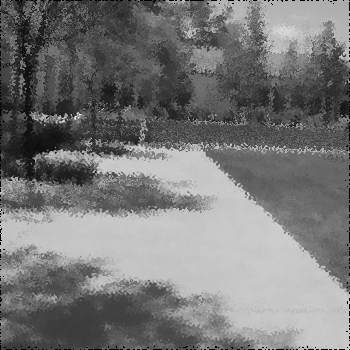}
		\parbox{.06\textwidth}{\centering ~}
        	}
	\centerline{
		\hfill
        \includegraphics[width = \figSize]{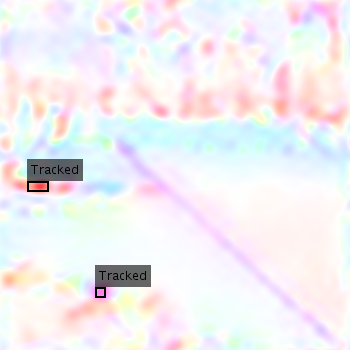}
		\hfill
        \includegraphics[width = \figSize]{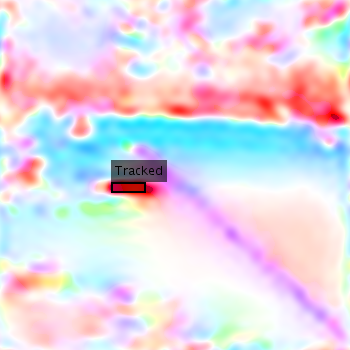}
		\hfill
		\includegraphics[width = \figSize]{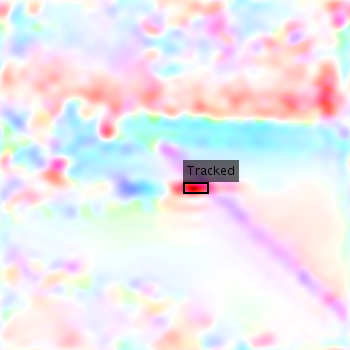}
		\hfill
		\includegraphics[width = \figSize]{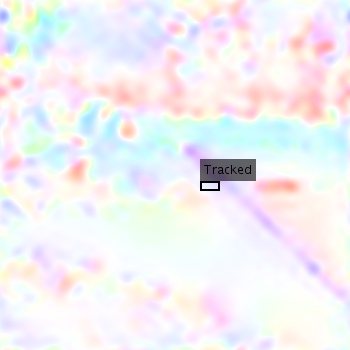}
		\hfill
		\includegraphics[width = \figSize]{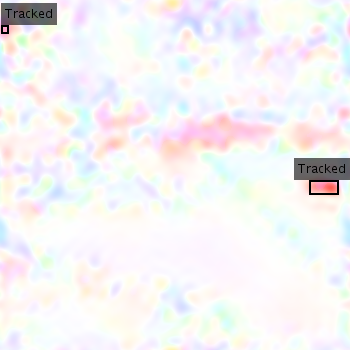}
		\hfill
		\includegraphics[width = \figSize]{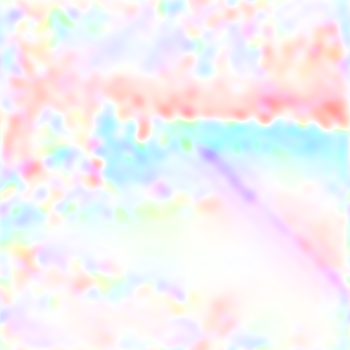}
		\parbox{.06\textwidth}{\centering \fbox{${\bf u}_\mathcal{C}$}}
        	}
        	
\end{minipage}}

\fbox{\begin{minipage}{.95\linewidth}
	\centerline{
		\hfill
        \includegraphics[width = \figSize]{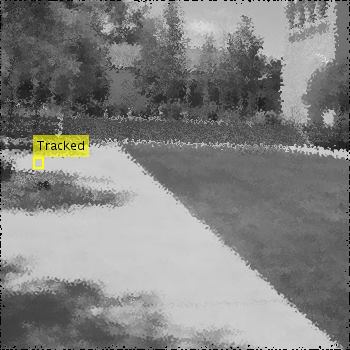}
		\hfill
        \includegraphics[width = \figSize]{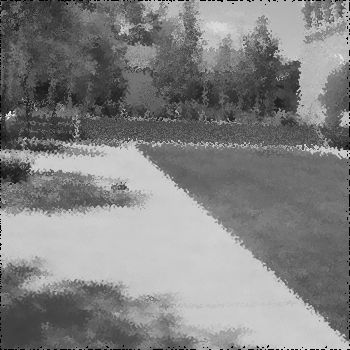}
		\hfill
		\includegraphics[width = \figSize]{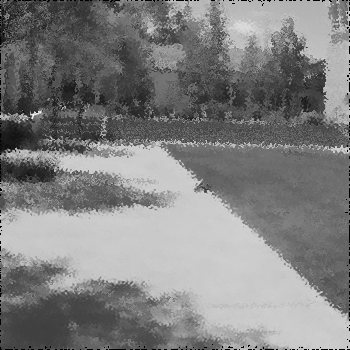}
		\hfill
		\includegraphics[width = \figSize]{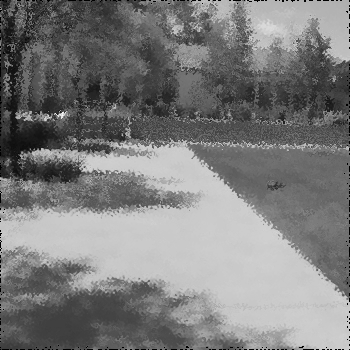}
		\hfill
		\includegraphics[width = \figSize]{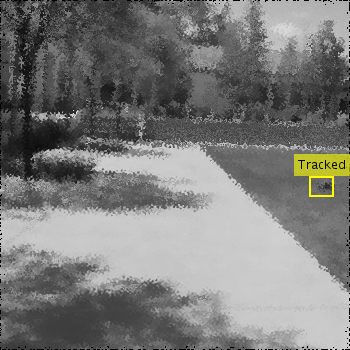}
		\hfill
		\includegraphics[width = \figSize]{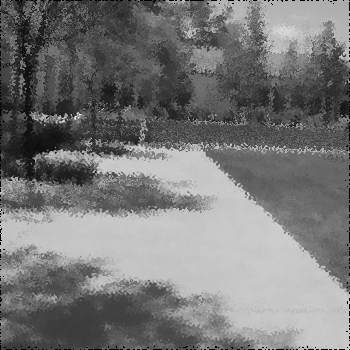}
		\parbox{.06\textwidth}{\centering ~}
        	}
	\centerline{
		\hfill
        \includegraphics[width = \figSize]{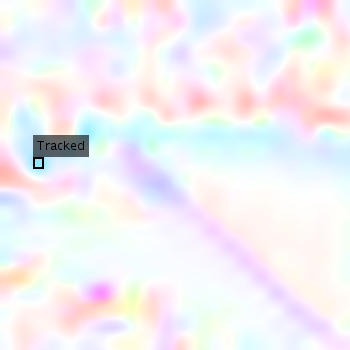}
		\hfill
        \includegraphics[width = \figSize]{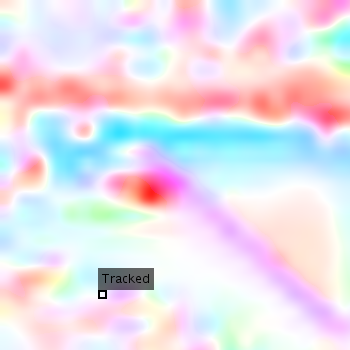}
		\hfill
		\includegraphics[width = \figSize]{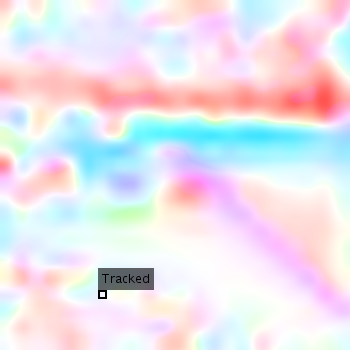}
		\hfill
		\includegraphics[width = \figSize]{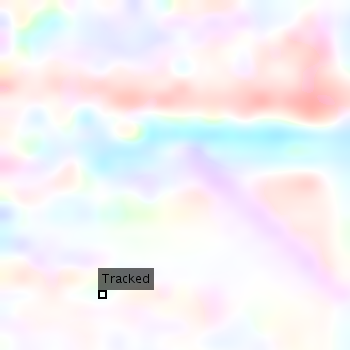}
		\hfill
		\includegraphics[width = \figSize]{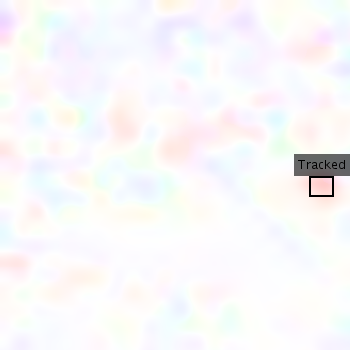}
		\hfill
		\includegraphics[width = \figSize]{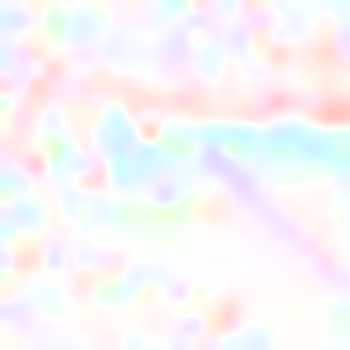}
		\parbox{.06\textwidth}{\centering \fbox{${\bf u}_\mathcal{W}$}}
        	}


\end{minipage}}

\fbox{\begin{minipage}{.95\linewidth}
	\centerline{
		\hfill
        \includegraphics[width = \figSize]{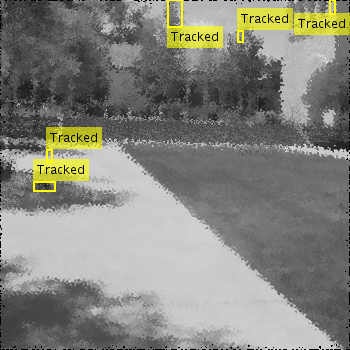}
		\hfill
        \includegraphics[width = \figSize]{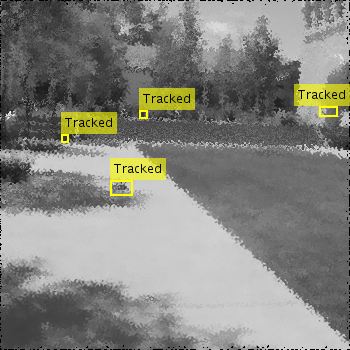}
		\hfill
		\includegraphics[width = \figSize]{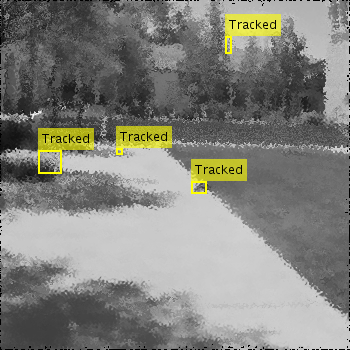}
		\hfill
		\includegraphics[width = \figSize]{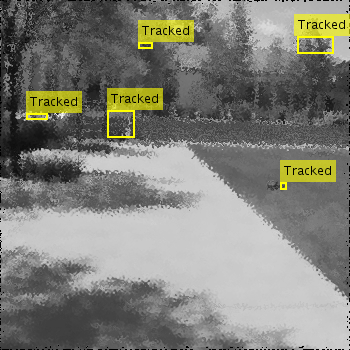}
		\hfill
		\includegraphics[width = \figSize]{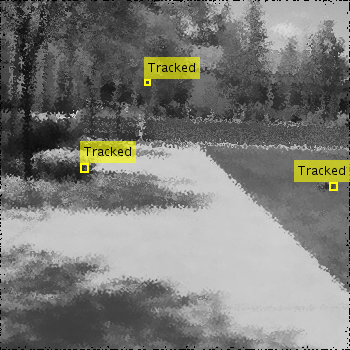}
		\hfill
		\includegraphics[width = \figSize]{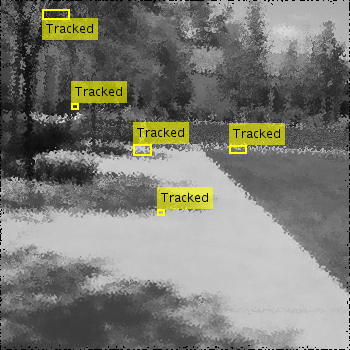}
		\parbox{.06\textwidth}{\centering ~}
        	}
	\centerline{
		\hfill
        \includegraphics[width = \figSize]{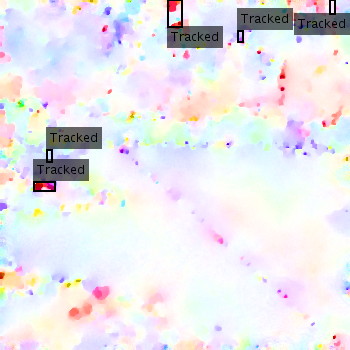}
		\hfill
        \includegraphics[width = \figSize]{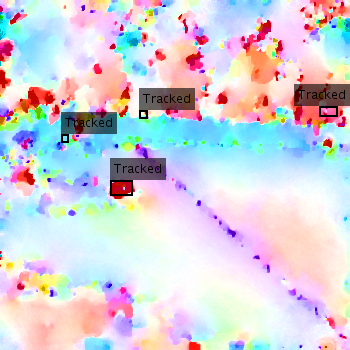}
		\hfill
		\includegraphics[width = \figSize]{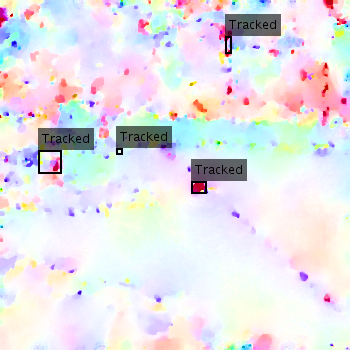}
		\hfill
		\includegraphics[width = \figSize]{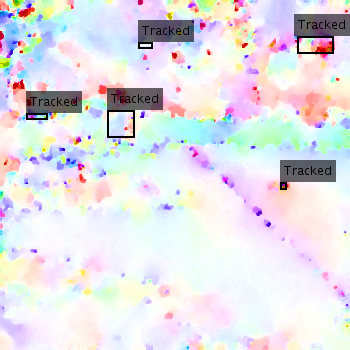}
		\hfill
		\includegraphics[width = \figSize]{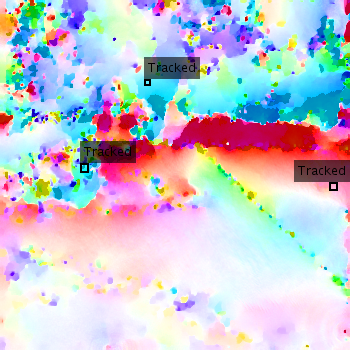}
		\hfill
		\includegraphics[width = \figSize]{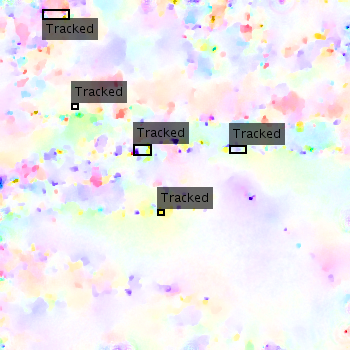}
		\parbox{.06\textwidth}{\centering \fbox{${\bf V}_c$}}
    }
\end{minipage}}
\small
\begin{minipage}{.9\linewidth}
\begin{center}
\begin{tabular}{ | c | c | c | c | c | c |}
\hline
Flow & F1$\uparrow$ & FDR$\downarrow$ & PPV$\uparrow$ & ACC$\uparrow$ & FNR$\downarrow$ \\ \hline
${\bf u}_\mathcal{C}$ 	&\bf  0.7351 &\bf  0.1920 &\bf  0.8080 &\bf  0.7188 & 0.0893 \\ \hline 
${\bf u}_\mathcal{W}$ 	& 0.3601 & 0.4911 & 0.5089 & 0.3571 & 0.1518 \\ \hline 
${\bf V}_{c}$ & 0.4408 & 0.6138 & 0.3862 & 0.3772 & \bf 0.0089 \\ \hline 
\end{tabular}
\end{center}
\end{minipage}
   \caption{Results from Courtyard2 which contains simulated atmospheric turbulence. Detection results are shown in image domain as well as a colorized version of the optical flow as to give context to the reader.}
\label{fig: Detection Courtyard 2}
\end{figure}
\begin{figure}[p!]
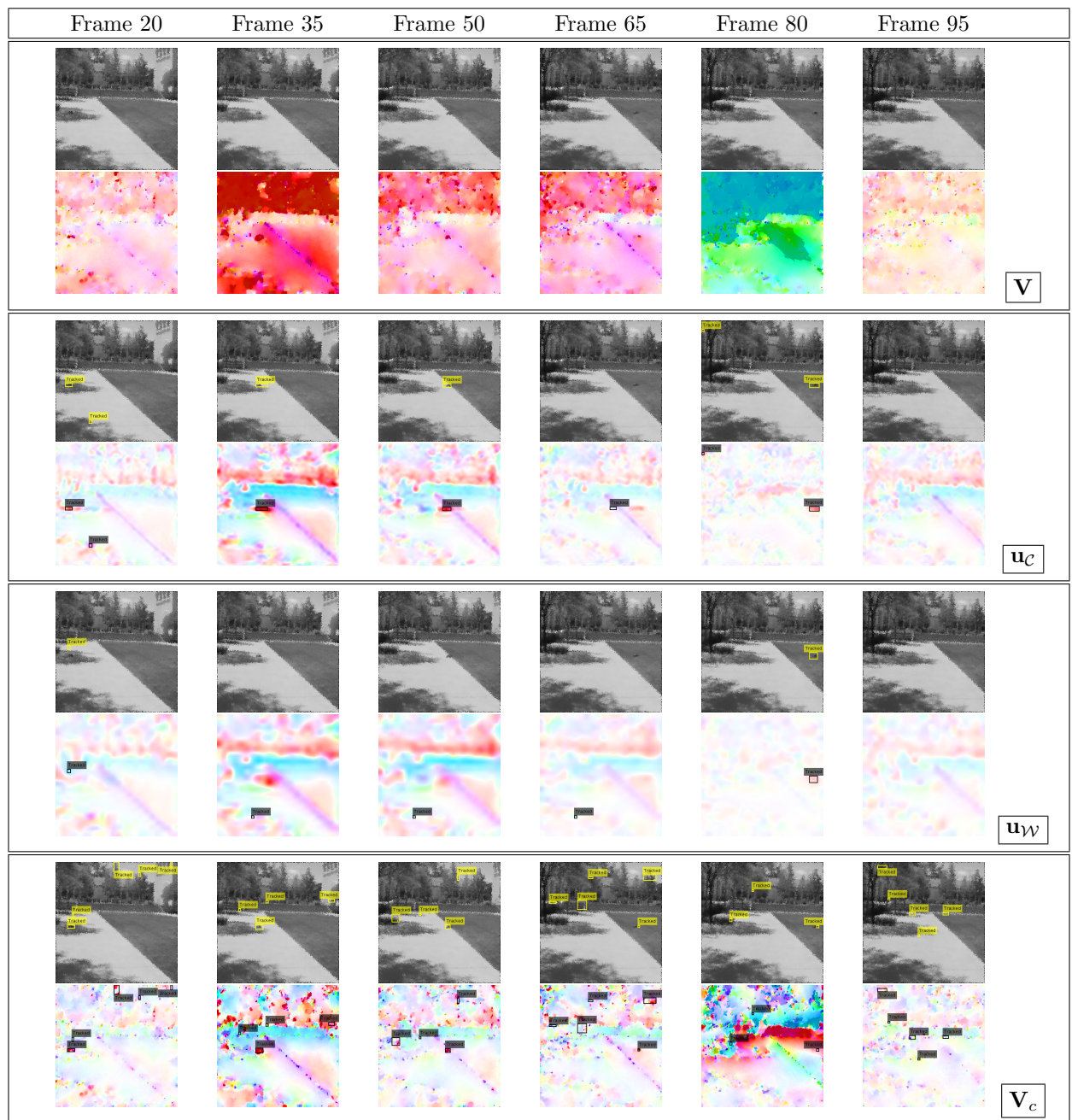

\centering
\fbox{\begin{minipage}{.95\linewidth}
\centerline{
\hfill
\parbox{\figSize}{\centering Frame 20}
\hfill
\parbox{\figSize}{\centering Frame 35}
\hfill
\parbox{\figSize}{\centering Frame 50}
\hfill
\parbox{\figSize}{\centering Frame 65}
\hfill
\parbox{\figSize}{\centering Frame 80}
\hfill
\parbox{\figSize}{\centering Frame 95}
\parbox{.06\textwidth}{\centering ~}
}   
\end{minipage}}
\fbox{\begin{minipage}{.95\linewidth}
	\centerline{
		\hfill
        \includegraphics[width = \figSize]{frame_1.png}
		\hfill
        \includegraphics[width = \figSize]{frame_2.png}
		\hfill
		\includegraphics[width = \figSize]{frame_3.png}
		\hfill
		\includegraphics[width = \figSize]{frame_4.png}
		\hfill
		\includegraphics[width = \figSize]{frame_5.png}
		\hfill
		\includegraphics[width = \figSize]{frame_6.png}
		\parbox{.06\textwidth}{\centering ~}
		}
	\centerline{
		\hfill
        \includegraphics[width = \figSize]{flow_1.png}
		\hfill
        \includegraphics[width = \figSize]{flow_2.png}
		\hfill
		\includegraphics[width = \figSize]{flow_3.png}
		\hfill
		\includegraphics[width = \figSize]{flow_4.png}
		\hfill
		\includegraphics[width = \figSize]{flow_5.png}
		\hfill
		\includegraphics[width = \figSize]{flow_6.png}
		\parbox{.06\textwidth}{\centering \fbox{${\bf V}$}}
		}
\end{minipage}}
\fbox{\begin{minipage}{.95\linewidth}
	\centerline{
		\hfill
        \includegraphics[width = \figSize]{detect_curv_1.png}
		\hfill
        \includegraphics[width = \figSize]{detect_curv_2.png}
		\hfill
		\includegraphics[width = \figSize]{detect_curv_3.png}
		\hfill
		\includegraphics[width = \figSize]{detect_curv_4.png}
		\hfill
		\includegraphics[width = \figSize]{detect_curv_5.png}
		\hfill
		\includegraphics[width = \figSize]{detect_curv_6.png}
		\parbox{.06\textwidth}{\centering ~}
        	}
	\centerline{
		\hfill
        \includegraphics[width = \figSize]{color_curv_1.png}
		\hfill
        \includegraphics[width = \figSize]{color_curv_2.png}
		\hfill
		\includegraphics[width = \figSize]{color_curv_3.png}
		\hfill
		\includegraphics[width = \figSize]{color_curv_4.png}
		\hfill
		\includegraphics[width = \figSize]{color_curv_5.png}
		\hfill
		\includegraphics[width = \figSize]{color_curv_6.png}
		\parbox{.06\textwidth}{\centering \fbox{${\bf u}_\mathcal{C}$}}
        	}
        	
\end{minipage}}

\fbox{\begin{minipage}{.95\linewidth}
	\centerline{
		\hfill
        \includegraphics[width = \figSize]{detect_wave_1.png}
		\hfill
        \includegraphics[width = \figSize]{detect_wave_2.png}
		\hfill
		\includegraphics[width = \figSize]{detect_wave_3.png}
		\hfill
		\includegraphics[width = \figSize]{detect_wave_4.png}
		\hfill
		\includegraphics[width = \figSize]{detect_wave_5.png}
		\hfill
		\includegraphics[width = \figSize]{detect_wave_6.png}
		\parbox{.06\textwidth}{\centering ~}
        	}
	\centerline{
		\hfill
        \includegraphics[width = \figSize]{color_wave_1.png}
		\hfill
        \includegraphics[width = \figSize]{color_wave_2.png}
		\hfill
		\includegraphics[width = \figSize]{color_wave_3.png}
		\hfill
		\includegraphics[width = \figSize]{color_wave_4.png}
		\hfill
		\includegraphics[width = \figSize]{color_wave_5.png}
		\hfill
		\includegraphics[width = \figSize]{color_wave_6.png}
		\parbox{.06\textwidth}{\centering \fbox{${\bf u}_\mathcal{W}$}}
        	}


\end{minipage}}

\fbox{\begin{minipage}{.95\linewidth}
	\centerline{
		\hfill
        \includegraphics[width = \figSize]{detect_fc_1.png}
		\hfill
        \includegraphics[width = \figSize]{detect_fc_2.png}
		\hfill
		\includegraphics[width = \figSize]{detect_fc_3.png}
		\hfill
		\includegraphics[width = \figSize]{detect_fc_4.png}
		\hfill
		\includegraphics[width = \figSize]{detect_fc_5.png}
		\hfill
		\includegraphics[width = \figSize]{detect_fc_6.png}
		\parbox{.06\textwidth}{\centering ~}
        	}
	\centerline{
		\hfill
        \includegraphics[width = \figSize]{color_fc_1.png}
		\hfill
        \includegraphics[width = \figSize]{color_fc_2.png}
		\hfill
		\includegraphics[width = \figSize]{color_fc_3.png}
		\hfill
		\includegraphics[width = \figSize]{color_fc_4.png}
		\hfill
		\includegraphics[width = \figSize]{color_fc_5.png}
		\hfill
		\includegraphics[width = \figSize]{color_fc_6.png}
		\parbox{.06\textwidth}{\centering \fbox{${\bf V}_c$}}
        	}
\end{minipage}}
\small

\begin{minipage}{.9\linewidth}
\begin{center}
\begin{tabular}{ | c | c | c | c | c | c |}
\hline
Flow & F1$\uparrow$ & FDR$\downarrow$ & PPV$\uparrow$ & ACC$\uparrow$ & FNR$\downarrow$ \\ \hline
${\bf u}_\mathcal{C}$ 	& \bf 0.5926 &\bf 0.1414 &\bf  0.8586 &\bf  0.5859 &  0.2727 \\ \hline 
${\bf u}_\mathcal{W}$ 	& 0.3973 & 0.2475 & 0.7525 & 0.3889 & 0.3636 \\ \hline 
${\bf V}_{c}$ & 0.4471 & 0.5047 & 0.4953 & 0.3741  & \bf 0.1212 \\ \hline 
\end{tabular}
\end{center}
\end{minipage}
\caption{Results from Courtyard4 which contains simulated atmospheric turbulence. Take note of the multiple false positive results in ${\bf V}_c$ from the simulated atmospheric turbulence.}
\end{figure}
\begin{figure}[p!]
\centering
\fbox{\begin{minipage}{.95\linewidth}
\centerline{
\hfill
\parbox{\figSize}{\centering Frame 20}
\hfill
\parbox{\figSize}{\centering Frame 35}
\hfill
\parbox{\figSize}{\centering Frame 50}
\hfill
\parbox{\figSize}{\centering Frame 65}
\hfill
\parbox{\figSize}{\centering Frame 80}
\hfill
\parbox{\figSize}{\centering Frame 95}
\parbox{.06\textwidth}{\centering ~}
}   
\end{minipage}}
\fbox{\begin{minipage}{.95\linewidth}
	\centerline{
		\hfill
        \includegraphics[width = \figSize]{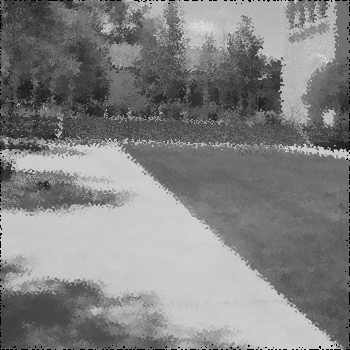}
		\hfill
        \includegraphics[width = \figSize]{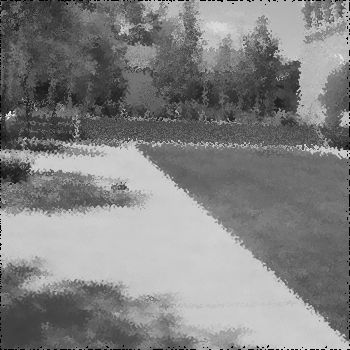}
		\hfill
		\includegraphics[width = \figSize]{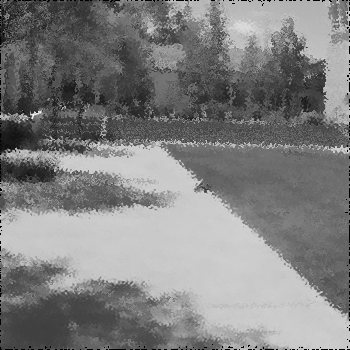}
		\hfill
		\includegraphics[width = \figSize]{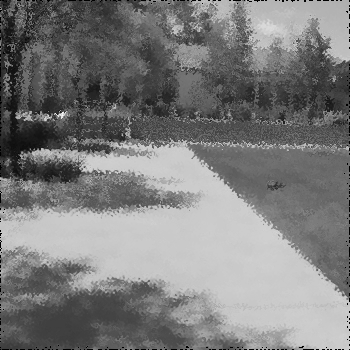}
		\hfill
		\includegraphics[width = \figSize]{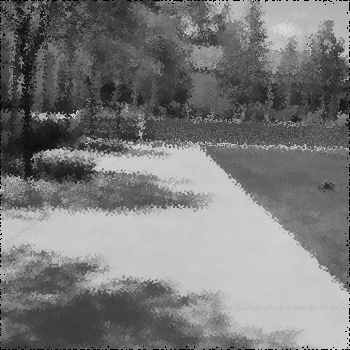}
		\hfill
		\includegraphics[width = \figSize]{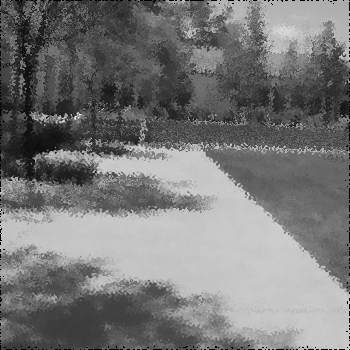}
		\parbox{.06\textwidth}{\centering ~}
		}
	\centerline{
		\hfill
        \includegraphics[width = \figSize]{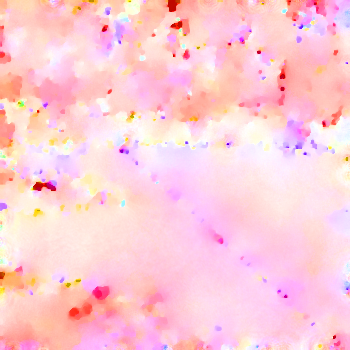}
		\hfill
        \includegraphics[width = \figSize]{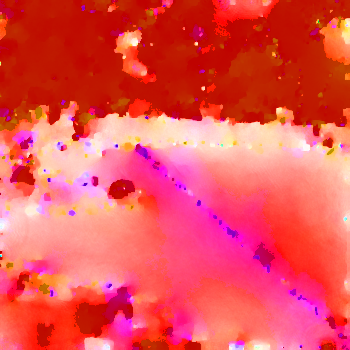}
		\hfill
		\includegraphics[width = \figSize]{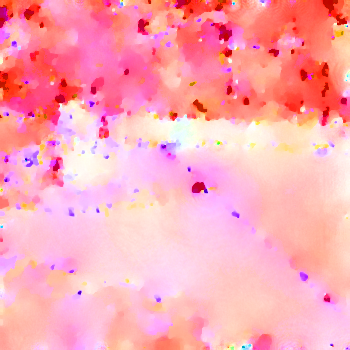}
		\hfill
		\includegraphics[width = \figSize]{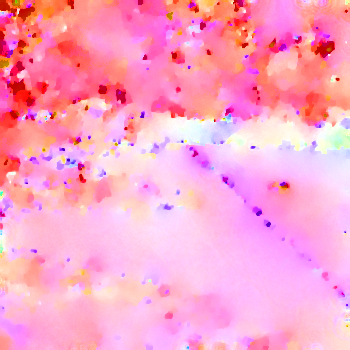}
		\hfill
		\includegraphics[width = \figSize]{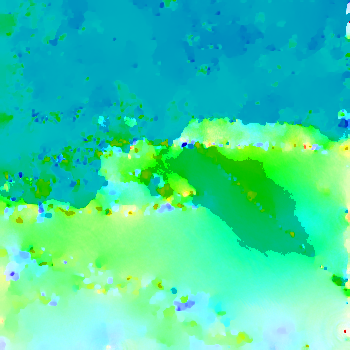}
		\hfill
		\includegraphics[width = \figSize]{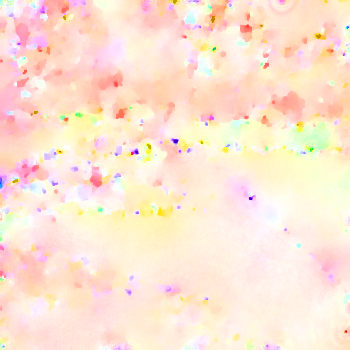}
		\parbox{.06\textwidth}{\centering \fbox{${\bf V}$}}
		}
\end{minipage}}
\fbox{\begin{minipage}{.95\linewidth}
	\centerline{
		\hfill
        \includegraphics[width = \figSize]{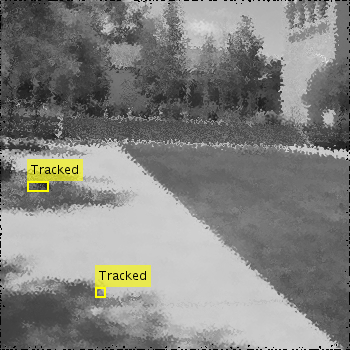}
		\hfill
        \includegraphics[width = \figSize]{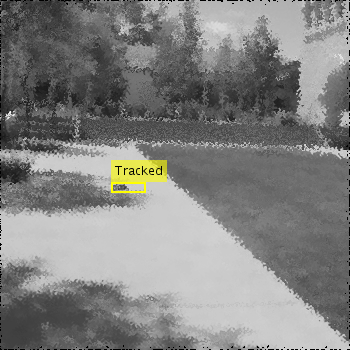}
		\hfill
		\includegraphics[width = \figSize]{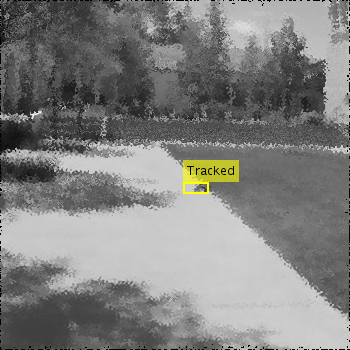}
		\hfill
		\includegraphics[width = \figSize]{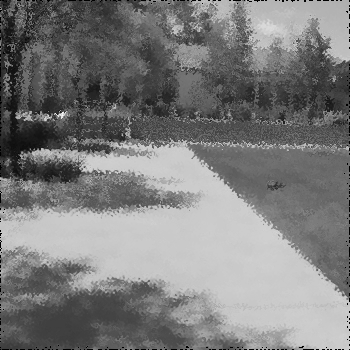}
		\hfill
		\includegraphics[width = \figSize]{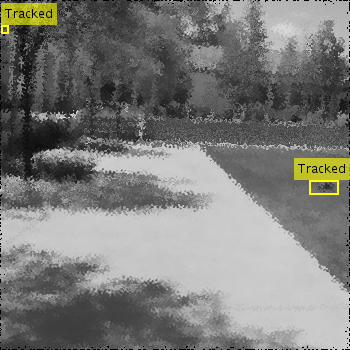}
		\hfill
		\includegraphics[width = \figSize]{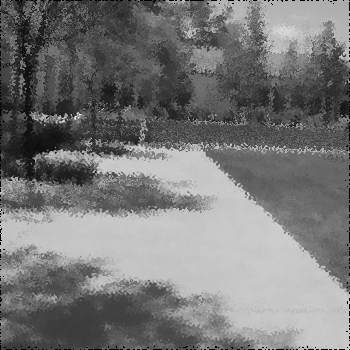}
		\parbox{.06\textwidth}{\centering ~}
        	}
	\centerline{
		\hfill
        \includegraphics[width = \figSize]{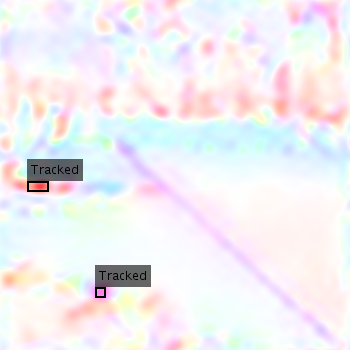}
		\hfill
        \includegraphics[width = \figSize]{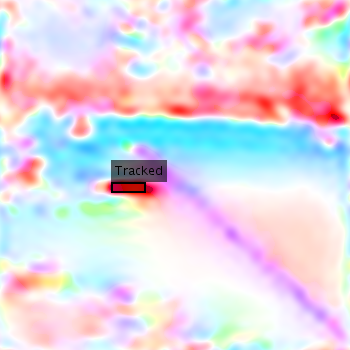}
		\hfill
		\includegraphics[width = \figSize]{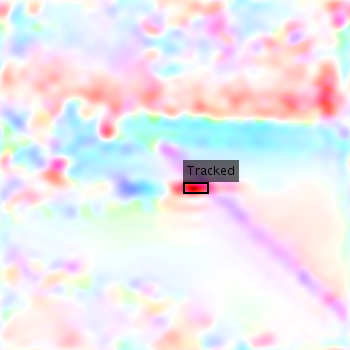}
		\hfill
		\includegraphics[width = \figSize]{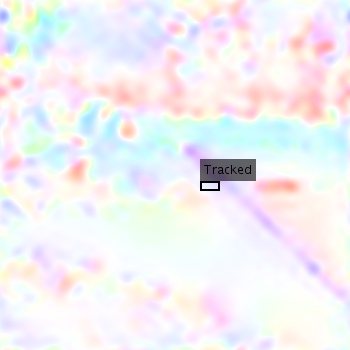}
		\hfill
		\includegraphics[width = \figSize]{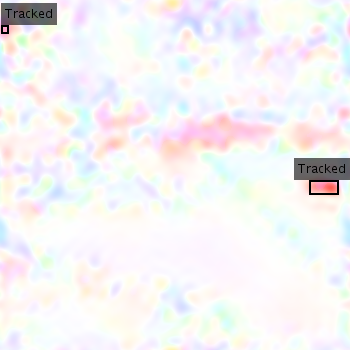}
		\hfill
		\includegraphics[width = \figSize]{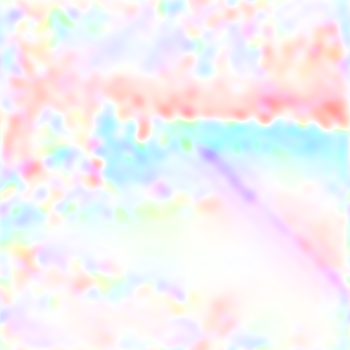}
		\parbox{.06\textwidth}{\centering \fbox{${\bf u}_\mathcal{C}$}}
        	}
        	
\end{minipage}}

\fbox{\begin{minipage}{.95\linewidth}
	\centerline{
		\hfill
        \includegraphics[width = \figSize]{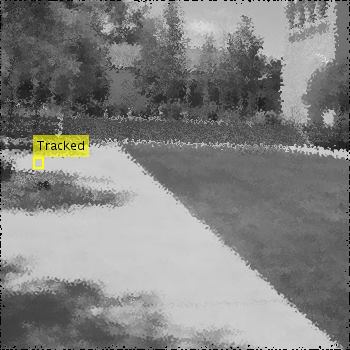}
		\hfill
        \includegraphics[width = \figSize]{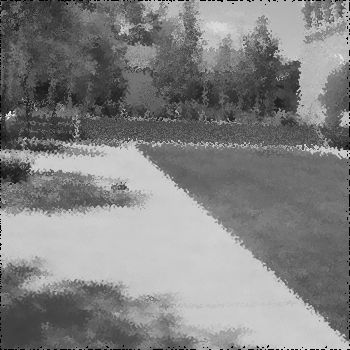}
		\hfill
		\includegraphics[width = \figSize]{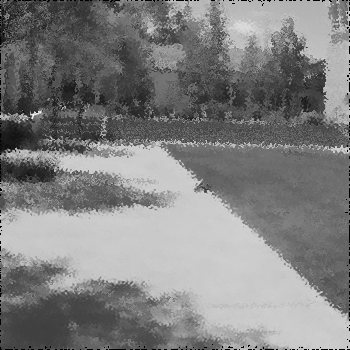}
		\hfill
		\includegraphics[width = \figSize]{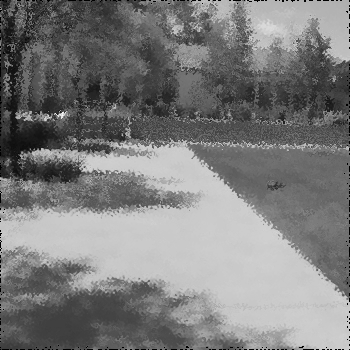}
		\hfill
		\includegraphics[width = \figSize]{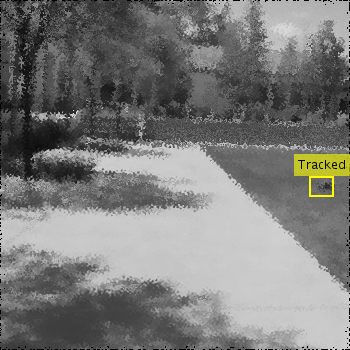}
		\hfill
		\includegraphics[width = \figSize]{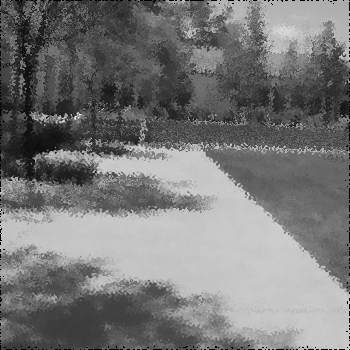}
		\parbox{.06\textwidth}{\centering ~}
        	}
	\centerline{
		\hfill
        \includegraphics[width = \figSize]{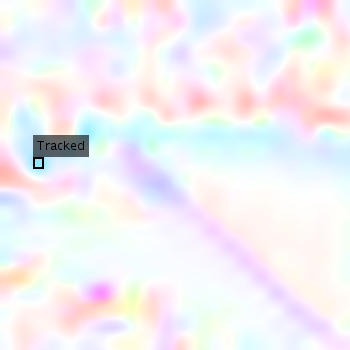}
		\hfill
        \includegraphics[width = \figSize]{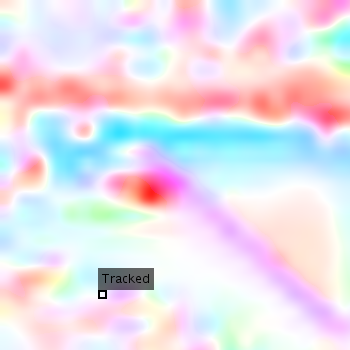}
		\hfill
		\includegraphics[width = \figSize]{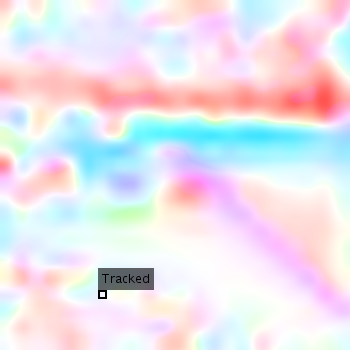}
		\hfill
		\includegraphics[width = \figSize]{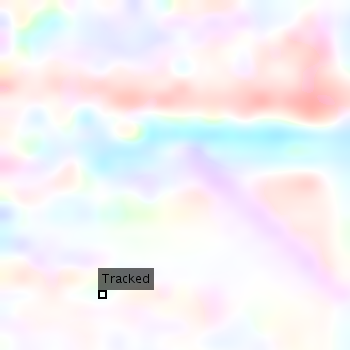}
		\hfill
		\includegraphics[width = \figSize]{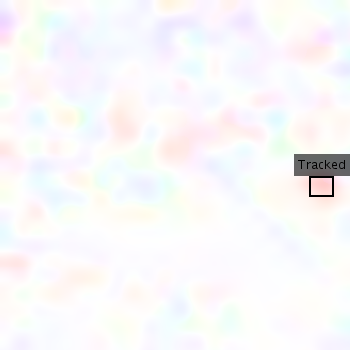}
		\hfill
		\includegraphics[width = \figSize]{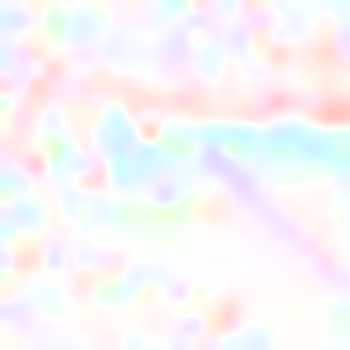}
		\parbox{.06\textwidth}{\centering \fbox{${\bf u}_\mathcal{W}$}}
        	}


\end{minipage}}

\fbox{\begin{minipage}{.95\linewidth}
	\centerline{
		\hfill
        \includegraphics[width = \figSize]{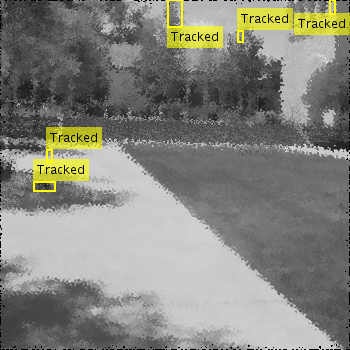}
		\hfill
        \includegraphics[width = \figSize]{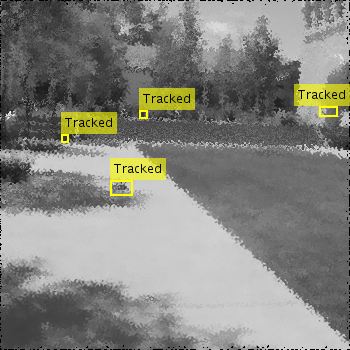}
		\hfill
		\includegraphics[width = \figSize]{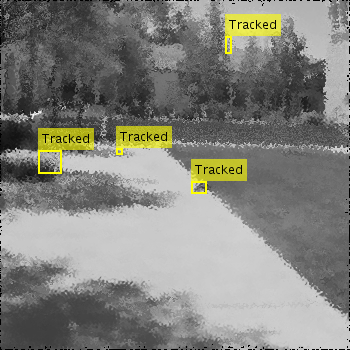}
		\hfill
		\includegraphics[width = \figSize]{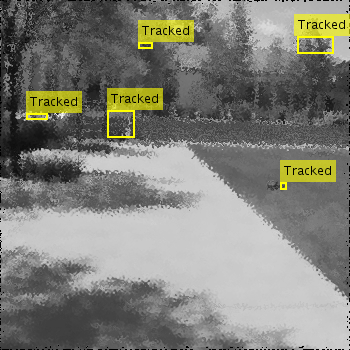}
		\hfill
		\includegraphics[width = \figSize]{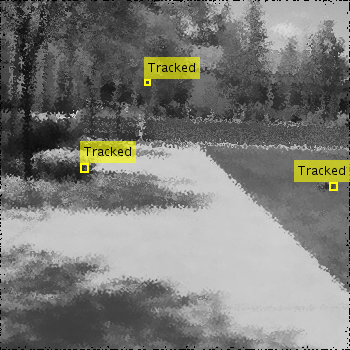}
		\hfill
		\includegraphics[width = \figSize]{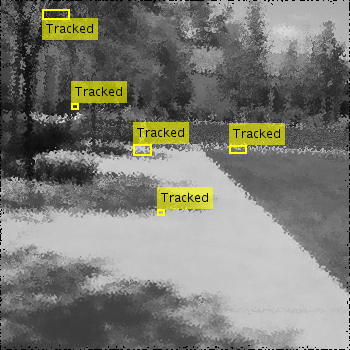}
		\parbox{.06\textwidth}{\centering ~}
        	}
	\centerline{
		\hfill
        \includegraphics[width = \figSize]{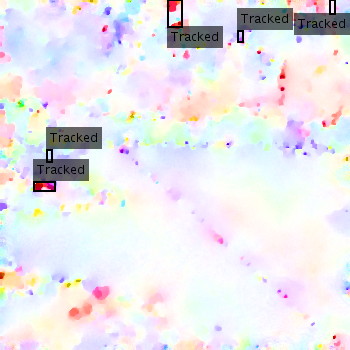}
		\hfill
        \includegraphics[width = \figSize]{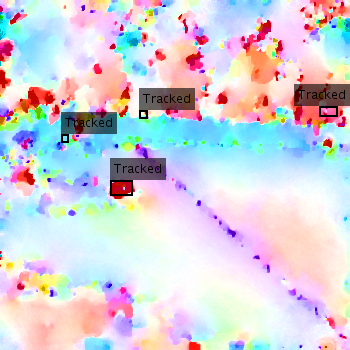}
		\hfill
		\includegraphics[width = \figSize]{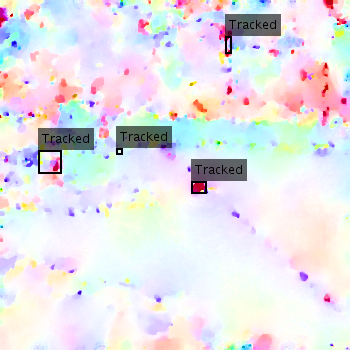}
		\hfill
		\includegraphics[width = \figSize]{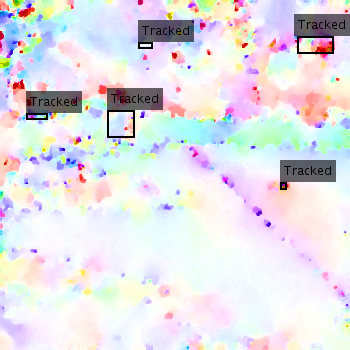}
		\hfill
		\includegraphics[width = \figSize]{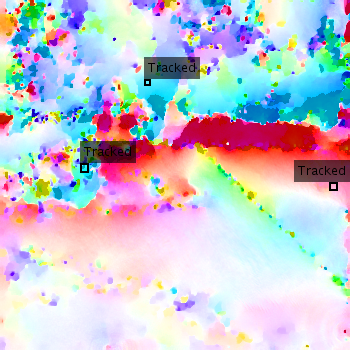}
		\hfill
		\includegraphics[width = \figSize]{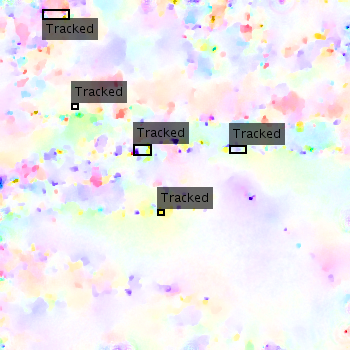}
		\parbox{.06\textwidth}{\centering \fbox{${\bf V}_c$}}
        	}
\end{minipage}}
\small

\begin{minipage}{.9\linewidth}
\begin{center}
\begin{tabular}{ | c | c | c | c | c | c |}
\hline
Flow & F1$\uparrow$ & FDR$\downarrow$ & PPV$\uparrow$ & ACC$\uparrow$ & FNR$\downarrow$ \\ \hline
${\bf u}_\mathcal{C}$ 	&\bf  0.7374 &\bf  0.1818 &\bf  0.8182 &\bf  0.7374 &0.0808 \\ \hline 
${\bf u}_\mathcal{W}$ 	& 0.2862 & 0.5000 & 0.5000 & 0.2778 & 0.2222 \\ \hline 
${\bf V}_{c}$ & 0.4158 & 0.5404 & 0.4596 & 0.3990 &\bf 0.0606 \\ \hline 
\end{tabular}
\end{center}
\end{minipage}
\caption{Results from Field1 which contains naturally forming atmospheric turbulence. Take note of the multiple false positive results in ${\bf V}_c$ from the naturally forming atmospheric turbulence. Secondarily, take note of the smearing in ${\bf u}_\mathcal{C}$ and ${\bf u}_\mathcal{W}$ due to the displacement of the object across the window over few frames due to camera motion.}
\label{fig: Detection Field1}
\end{figure}

\section{Conclusion}
\label{chap: Conclusion}
\begin{figure}[t]
\centering
\includegraphics[width=.98\linewidth]{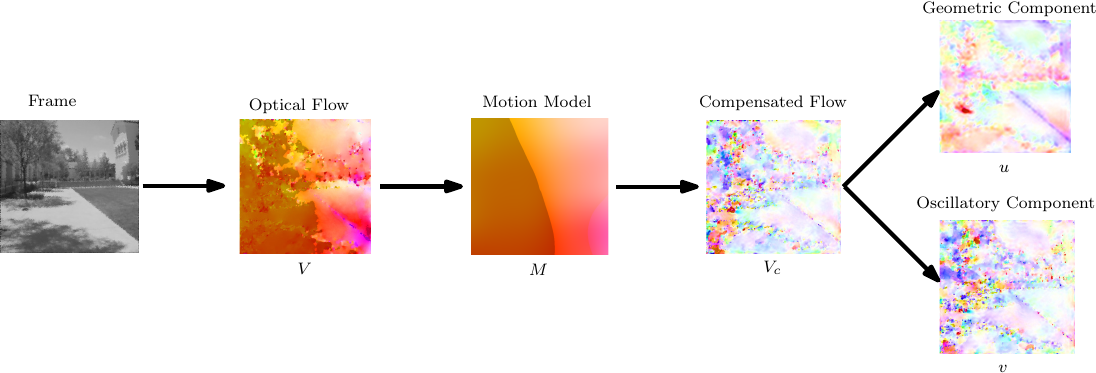}
\caption{The decomposition process for the simulated turbulence sequence \textit{Courtyard4} at frame 30 when using the empirical camera motion model and curvelet decomposition.}
\label{fig: Frames 4}
\end{figure}
In this study we have discussed the creation of an image set that contains global camera motion with both simulated or natural atmospheric turbulence as well as the development of an algorithm that detects moving objects from the procured image set. 
The demonstrated algorithm is able to take an input data sequence, determine its optical flow, and decompose it into its camera motion, geometric, and oscillatory flow fields as summarized in Figure~\ref{fig: Frames 4}.  Next, we perform a detection and tracking method on the geometric flow field in order to determine locations within the image set where moving objects exist.
The detection of these objects, as well as the preprocessing steps taken to extract the camera motion flow, provide a novel solution to a previously understudied problem.
In future work, we will address extensions in four fronts: the inclusion of depth in the motion compensation, the runtime of the cartoon+texture decomposition, the inclusion of camera motion in the Kalman filter prediction step, and finally the confirmation on more datasets.\\

In the model for global motion flow we have assumed a smooth function to describe the change in $\mathcal{Z}$.
In reality, this may not be a fair assumption as a change in depth can be instantaneous.
\begin{figure}[b]
\centering
\hfill
\includegraphics[width=.3\textwidth]{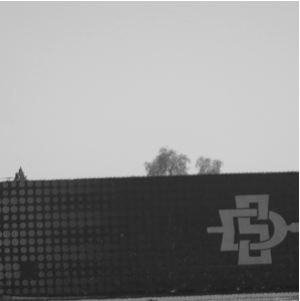}
\hfill
\includegraphics[width=.3\textwidth]{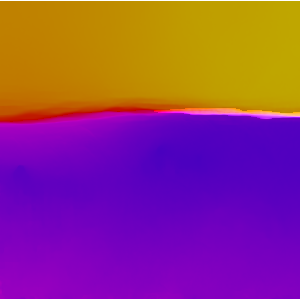}
\hfill
\includegraphics[width=.3\textwidth]{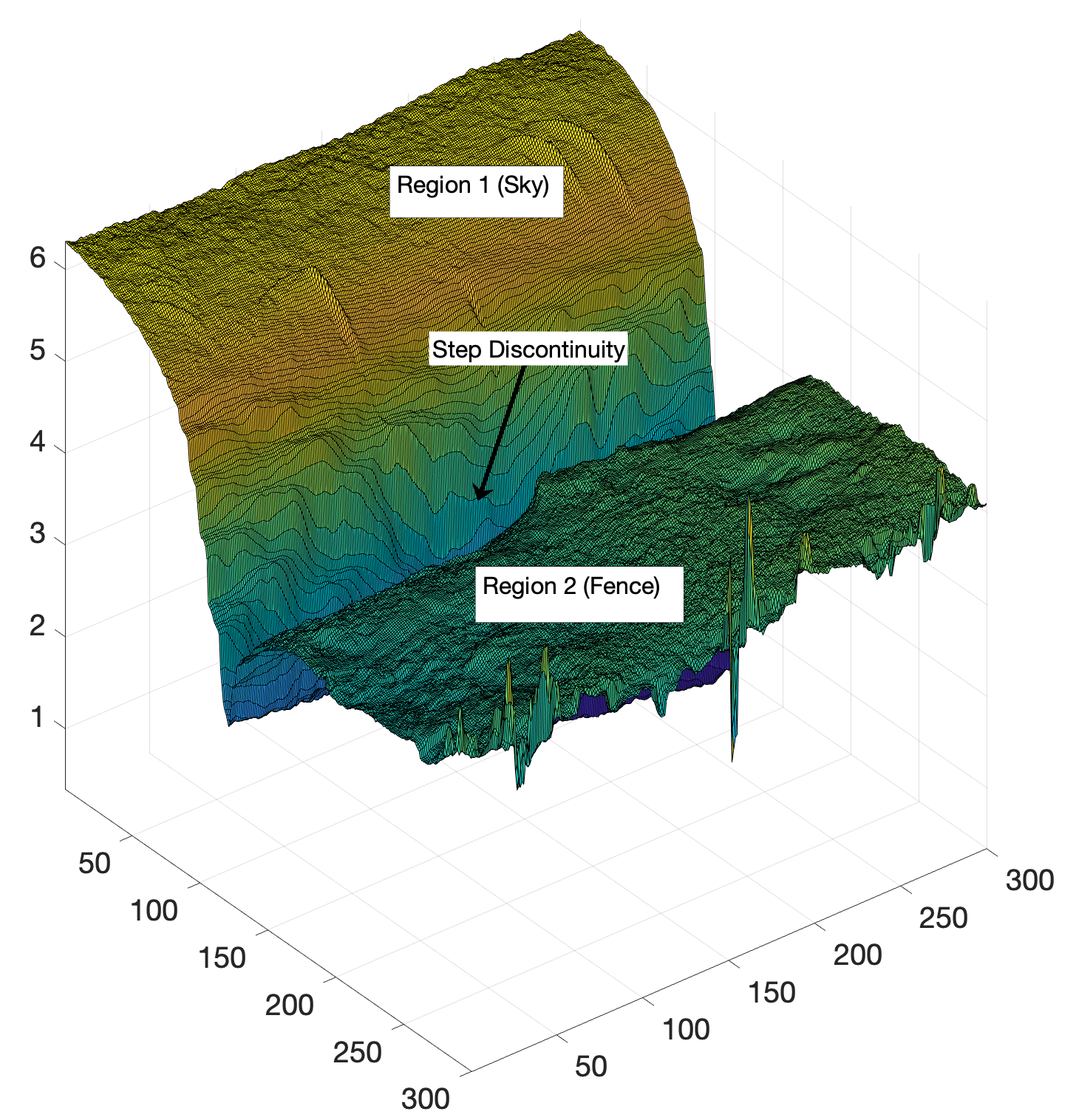}
\hfill
\caption{Depiction of the step discontinuity in the magnitude of the optical flow corresponding to an instantaneous change in depth $\mathcal{Z}$. Frame 90 from Field4 (left) is presented with a colorized depiction of the optical flow (center) and the magnitude of the optical flow (right).}
\label{fig: Step Dis}
\end{figure}
From the camera's perspective, an object that is occluding the field of view \textit{i.e.} a mountain range or a building, may introduce such phenomena.
This sharp change will cause a step discontinuity in the optical flow components.
In order to handle this step discontinuity, we require a method that segments the flow into regions corresponding to each region of depth.
An example of this step discontinuity is shown in Figure~\ref{fig: Step Dis} where the fence in the sequence causes an instantaneous change in the depth of the image field of view.
Once the regions are segmented, the empirical Gaussian smoothing method will provide a motion model for each region.\\

Recalling Figure~\ref{fig: CurveletWavelet Runtime} we saw the curvelet decomposition did not scale well to larger video sizes.
This is a shortcoming that renders real time application impossible.
We saw that a naive implementation of the wavelet transform provides acceptable results but failed to capture detailed temporal information.
This is certainly due to the bias from scaling the spatial domain equally with the temporal domain.
With a sequence with only a few frames, we have rich temporal information over small scales, while a larger video size has no need for such detailed information.
Therefore an implementation of a two-dimensional wavelet transform over the spatial information and a one-dimensional transformation over the temporal information is suggested.
This method should preserve the wavelet runtime, while providing a fair weighting between the scales for the problem.
This approach would be also lend well to coupling sequential data sequences.
The current decomposition process operates on three-dimensional data cubes, making coupling another data set difficult to implement.
By structuring the decomposition in a way that is separating the spatial and temporal information, it motivates the idea that we may couple data by only taking a new spatial wavelet transform and adding the new temporal information from a new frame.\\

When using the detection and tracking algorithm, we notice that, even with the inclusion of false positive detections, the algorithm is able to detect and track the moving object accurately.
During the implementation of the Kalman Filter, we provide simply a constant acceleration motion model, without the input of the model for camera motion.
Without knowledge of camera motion, the Kalman filter is not working as efficiently as possible as its predictions are impacted by the camera motion.
As we have already built a motion model, we have a prediction for the location of each pixel in the next frame.
If we were to couple this knowledge with the Kalman filter prediction step, a more accurate tracking algorithm could be created.

Finally, this work only contained a limited data set, some of which did not contain heavy enough atmospheric turbulence to offset the speed at which the object was moving.
As we detect the object by adaptive thresholding of the speed of the flow, we expect that a fast moving object will be detected in most situations.
A dataset that contains a slow moving object with a moving camera would be a great demonstration and further confirmation of this promising algorithm.

\section{Acknowledgment}
This work was supported by the Air Force Office of Scientific Research under the grant number FA9550-15-1-0065.

\appendix

\section{Camera motion model derivation}
\label{sec: Motion Model}

We previously observed that the flow induced by camera motion is a smooth function by appearance (see Figure~\ref{fig:Camera Motion Motivation 1}).
One model for the optical flow induced by global camera motion is derived from the pinhole camera model \eqref{eqn: Pinhole Camera} and has been previously presented by Thompson and Pong \cite{thompson1990detecting} as well as Trucco and Verri \cite{trucco1998introductory}.
From Trucco and Verri, the pinhole camera model is given by~\eqref{eqn: Pinhole Camera}
\begin{equation}
{\bf p} = \foc\frac{{\bf \mathcal{P}}}{\mathcal{Z}}
\label{eqn: Pinhole Camera}
\end{equation}
where ${\bf \mathcal{P}} = \left(\mathcal{X}, \mathcal{Y}, \mathcal{Z}\right)$ in the usual 3D camera reference basis.
The projection center is chosen to be the origin of that reference basis and $\foc$ denotes the focal length.
This model provides a relation between points in the image scene $\mathcal{P}$ and their projections in the image plane ${\bf p}=[x, y, \foc]^{\top}$.
As the image plane is at a constant location along the optical axis, the third coordinate $\foc$ is dropped from the notation to have the point in the image plane ${\bf p}=[x, y]^{\top}$.
A depiction of this projection, from Vismara~{\cite{vismara2015monitoring} is shown in Figure~\ref{fig: Pinhole Camera}.
The relative motion between 
${\bf \mathcal{P}}$ and the camera is described as
\begin{equation}
{\bf \mathcal{V}} = -{\bf T} -\omega\times {\bf \mathcal{P}},
\label{eqn: Real world velocity}
\end{equation}
where ${\bf T}=(T_x,T_y,T_z)$ is the translation component of motion and $\omega=(\omega_x,\omega_y,\omega_z)$ the angular component which are constant at a given frame for all arbitrary points $\bf \mathcal{P}$ as we have rigid body camera motion.
We denote $\bf P, Y, R$ the optical flows induced by pitch (\textit{i.e.} by $\omega_x$), yaw (\textit{i.e.} by $\omega_y$) and roll (\textit{i.e.} by $\omega_z$).
Then \eqref{eqn: Real world velocity} expressed by its coordinates becomes,
\begin{align}
\mathcal{V}_x &= -T_x - \omega_y \mathcal{Z} + \omega_z \mathcal{Y} \nonumber \\
\mathcal{V}_y &= -T_y - \omega_z \mathcal{X} + \omega_x \mathcal{Z} \nonumber \\
\mathcal{V}_z &= -T_z - \omega_x \mathcal{Y} + \omega_y \mathcal{X}.
\label{eqn: Velocity Components}
\end{align}
\begin{figure}[t]
\centering
\includegraphics[width=.6\textwidth]{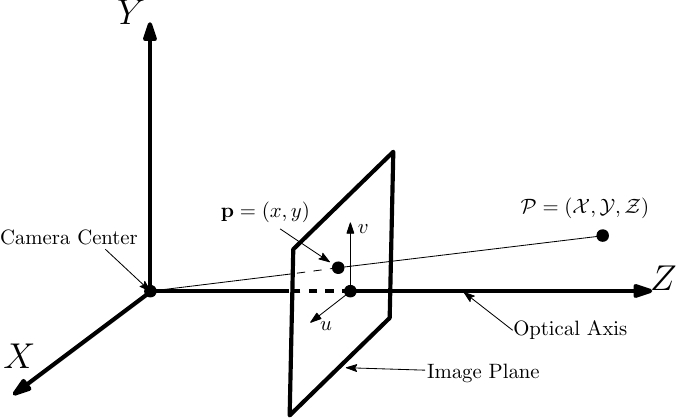}
\caption{Illustration of the pinhole camera model having the focal plane in front of the camera center. The image point $\bf p$ is located at $\foc$ on the optical axis $Z$.}
\label{fig: Pinhole Camera}
\end{figure}
To obtain the relation between the velocity of $\mathcal{P}$ in real-world coordinates and the corresponding velocity of $\bf p$ on the image plane, we take the time derivative of both sides of equation \eqref{eqn: Pinhole Camera} to obtain through the quotient rule our predicted optical flow,
\begin{equation}
{\bf V} = \foc\frac{\mathcal{Z}{\bf \mathcal{V}} - \mathcal{V}_z {\bf \mathcal{P}}}{\mathcal{Z}^2},
\label{eqn: Image to Real world velocity}
\end{equation}
where by substituting \eqref{eqn: Real world velocity} and \eqref{eqn: Velocity Components} into \eqref{eqn: Image to Real world velocity} we get
\begin{align}
V_{x} &= \foc\frac{\mathcal{Z}\mathcal{V}_x - \mathcal{V}_z\mathcal{X}}{Z^2} \nonumber \\
V_{y} &= \foc\frac{\mathcal{Z}\mathcal{V}_y - \mathcal{V}_zY}{\mathcal{Z}^2}.
\end{align}
Converting from real world coordinates to 2D image plane coordinates, we get
\begin{align}
V_x &= \foc\frac{\mathcal{Z}\mathcal{V}_x - V_z\mathcal{Z}\frac{x}{\foc}}{\mathcal{Z}^2} = \foc\frac{\mathcal{V}_x - \mathcal{V}_z\frac{x}{\foc}}{\mathcal{Z}}\nonumber \\
V_y &= \foc\frac{\mathcal{Z}\mathcal{V}_y - V_z\mathcal{Z}\frac{y}{\foc}}{\mathcal{Z}^2} = \foc\frac{\mathcal{V}_y - \mathcal{V}_z\frac{y}{\foc}}{\mathcal{Z}}
\label{eqn: pixel velocity}
\end{align}
and expanding $\mathcal{V}_x$, $\mathcal{V}_y$, $\mathcal{V}_z$ into 2D pixel coordinates, we get
\begin{align}
\mathcal{V}_x &= -T_x - \omega_y \mathcal{Z} + \omega_z \frac{y}{\foc} \nonumber \\
\mathcal{V}_y &= -T_y - \omega_z \frac{x}{\foc} + \omega_x \mathcal{Z} \nonumber \\
\mathcal{V}_z &= -T_z - \omega_x \frac{y}{\foc} + \omega_y \frac{x}{\foc},
\label{eqn: pixel Velocity Components}
\end{align}
then by combining \eqref{eqn: pixel velocity} and \eqref{eqn: pixel Velocity Components} we finally obtain,
\begin{align}
V_x &= \frac{T_z x - T_x \foc}{\mathcal{Z}} + \omega_x\frac{xy}{\foc} - \omega_y\left( \foc + \frac{x^2}{\foc} \right) + \omega_z y \nonumber \\
V_y &= \frac{T_z y - T_y \foc}{\mathcal{Z}} + \omega_x\left( \foc + \frac{y^2}{\foc} \right) - \omega_y\frac{xy}{\foc} - \omega_z x.
\label{eqn: Image to Real world velocity components}
\end{align}
We observe that the velocity field is the sum of two components, one containing information about camera translation and the other containing information about rotation. 
Hence we denote the translation components to be
\begin{align}
V_x^T &= \frac{T_z x - T_x \foc}{\mathcal{Z}} \nonumber \\
V_y^T &= \frac{T_z y - T_y \foc}{\mathcal{Z}},
\end{align}
and the rotation components to be,
\begin{align}
V_x^\omega &= \omega_x\frac{xy}{\foc} - \omega_y\left( \foc + \frac{x^2}{\foc} \right) + \omega_z y \nonumber \\
V_y^\omega &= \omega_x\left( \foc + \frac{y^2}{\foc} \right) - \omega_y\frac{xy}{\foc} - \omega_z x.
\label{eqn: Motion model}
\end{align}
We notice that in each velocity component, information on depth $\mathcal{Z}$ and rotation $\omega$ are decoupled. This shows that the part of the velocity field that depends 
on angular velocity does not carry information on depth.

\bibliographystyle{spiebib} 
\bibliography{report} 

\begin{thebibliography}{10}

\bibitem{gilles2018detection}
Gilles, J., Alvarez, F., Ferrante, N., Fortman, M., Tahir, L., Tarter, A., and von Seeger, A., ``Detection of moving objects through turbulent media. decomposition of oscillatory vs non-oscillatory spatio-temporal vector fields,'' {\em Image and Vision Computing}~{\bf 73},  40--55 (2018).

\bibitem{shaikh2014moving}
Shaikh, S.~H., Saeed, K., and Chaki, N., ``Moving object detection using background subtraction,'' in [{\em Moving Object Detection Using Background Subtraction}{\nolinebreak\hspace{0.1em}]},   15--23, Springer (2014).

\bibitem{yi2010moving}
Yi, Z. and Liangzhong, F., ``Moving object detection based on running average background and temporal difference,'' in [{\em Intelligent Systems and Knowledge Engineering (ISKE), 2010 International Conference on}{\nolinebreak\hspace{0.1em}]},   270--272, IEEE (2010).

\bibitem{chauhan2013moving}
Chauhan, A.~K. and Krishan, P., ``Moving object tracking using gaussian mixture model and optical flow,'' {\em International Journal of Advanced Research in Computer Science and Software Engineering}~{\bf 3}(4) (2013).

\bibitem{lecun2015deep}
LeCun, Y., Bengio, Y., and Hinton, G., ``Deep learning,'' {\em nature}~{\bf 521}(7553),  436 (2015).

\bibitem{patel2013moving}
Patel, H.~A. and Thakore, D.~G., ``Moving object tracking using {Kalman} filter,'' {\em International Journal of Computer Science and Mobile Computing}~{\bf 2}(4),  326--332 (2013).

\bibitem{gordon2004beyond}
Gordon, N., Ristic, B., and Arulampalam, S., ``Beyond the {Kalman} filter: Particle filters for tracking applications,'' {\em Artech House, London}~{\bf 830},  5 (2004).

\bibitem{briechle2001template}
Briechle, K. and Hanebeck, U.~D., ``Template matching using fast normalized cross correlation,'' in [{\em Optical Pattern Recognition XII}{\nolinebreak\hspace{0.1em}]},   {\bf 4387},  95--103, International Society for Optics and Photonics (2001).

\bibitem{comaniciu2000real}
Comaniciu, D., Ramesh, V., and Meer, P., ``Real-time tracking of non-rigid objects using mean shift,'' in [{\em Proceedings IEEE Conference on Computer Vision and Pattern Recognition. CVPR 2000 (Cat. No. PR00662)}{\nolinebreak\hspace{0.1em}]},   {\bf 2},  142--149, IEEE (2000).

\bibitem{avidan2001support}
Avidan, S., ``Support vector tracking,'' in [{\em Proceedings of the 2001 IEEE Computer Society Conference on Computer Vision and Pattern Recognition. CVPR 2001}{\nolinebreak\hspace{0.1em}]},   {\bf 1},  I--I, IEEE (2001).

\bibitem{zhou2003background}
Zhou, Y. and Tao, H., ``A background layer model for object tracking through occlusion,'' in [{\em Proceedings Ninth IEEE International Conference on Computer Vision}{\nolinebreak\hspace{0.1em}]},   1079--1085, IEEE (2003).

\bibitem{rosenhahn2006comparison}
Rosenhahn, B., Brox, T., Cremers, D., and Seidel, H.-P., ``A comparison of shape matching methods for contour based pose estimation,'' in [{\em International Workshop on Combinatorial Image Analysis}{\nolinebreak\hspace{0.1em}]},   263--276, Springer (2006).

\bibitem{balaji2017survey}
Balaji, S. and Karthikeyan, S., ``A survey on moving object tracking using image processing,'' in [{\em 2017 11th international conference on intelligent systems and control (ISCO)}{\nolinebreak\hspace{0.1em}]},   469--474, IEEE (2017).

\bibitem{yazdi2018new}
Yazdi, M. and Bouwmans, T., ``New trends on moving object detection in video images captured by a moving camera: A survey,'' {\em Computer Science Review}~{\bf 28},  157--177 (2018).

\bibitem{giancoli2008physics}
Giancoli, D.~C.,  [{\em Physics for scientists \& engineers with modern physics}{\nolinebreak\hspace{0.1em}]}, vol.~2, Pearson Education (2008).

\bibitem{hyde2010determining}
Hyde, M.~W., Schmidt, J.~D., Havrilla, M.~J., and Cain, S.~C., ``Determining the complex index of refraction of an unknown object using turbulence-degraded polarimetric imagery,'' Tech. Rep.~12 (2010).

\bibitem{gilles2017open}
Gilles, J. and Ferrante, N.~B., ``Open turbulent image set ({OTIS}),'' {\em Pattern Recognition Letters}~{\bf 86},  38--41 (2017).

\bibitem{tahtali2005restoration}
Tahtali, M., Fraser, D., and Lambert, A., ``Restoration of non-uniformly warped images using a typical frame as prototype,'' in [{\em TENCON 2005 2005 IEEE Region 10}{\nolinebreak\hspace{0.1em}]},   1--6, IEEE (2005).

\bibitem{horn1981determining}
Horn, B.~K. and Schunck, B.~G., ``Determining optical flow,'' {\em Artificial intelligence}~{\bf 17}(1-3),  185--203 (1981).

\bibitem{zach2007duality}
Zach, C., Pock, T., and Bischof, H., ``A duality based approach for realtime {TV-L1} optical flow,'' in [{\em Joint Pattern Recognition Symposium}{\nolinebreak\hspace{0.1em}]},   214--223, Springer (2007).

\bibitem{trucco1998introductory}
Trucco, E. and Verri, A.,  [{\em Introductory techniques for 3-D computer vision}{\nolinebreak\hspace{0.1em}]}, vol.~201, Prentice Hall Englewood Cliffs (1998).

\bibitem{Candes2005}
Cand\`es, E.~J., Demanet, L., Donoho, D.~L., and Ying, L., ``Fast discrete curvelet transforms,'' {\em Multiscale Modeling and Simulation}~{\bf 5}(3),  861--899 (2005).

\bibitem{kohavi1995study}
Kohavi, R. et~al., ``A study of cross-validation and bootstrap for accuracy estimation and model selection,'' in [{\em Ijcai}{\nolinebreak\hspace{0.1em}]},   {\bf 14}(2),  1137--1145, Montreal, Canada (1995).

\bibitem{kalman1960new}
Kalman, R.~E., ``A new approach to linear filtering and prediction problems,'' {\em Journal of basic Engineering}~{\bf 82}(1),  35--45 (1960).

\bibitem{munkres1957algorithms}
Munkres, J., ``Algorithms for the assignment and transportation problems,'' {\em Journal of the society for industrial and applied mathematics}~{\bf 5}(1),  32--38 (1957).

\bibitem{miller1997optimizing}
Miller, M.~L., Stone, H.~S., and Cox, I.~J., ``Optimizing murty's ranked assignment method,'' {\em IEEE Transactions on Aerospace and Electronic Systems}~{\bf 33}(3),  851--862 (1997).

\bibitem{fawcett2006introduction}
Fawcett, T., ``An introduction to roc analysis,'' {\em Pattern recognition letters}~{\bf 27}(8),  861--874 (2006).

\bibitem{thompson1990detecting}
Thompson, W.~B. and Pong, T.-C., ``Detecting moving objects,'' {\em International journal of computer vision}~{\bf 4}(1),  39--57 (1990).

\bibitem{vismara2015monitoring}
VISMARA, C., ``Monitoring human state in a robotic assistive platform: data acquisition and person detection systems,'' (2015).

\end{thebibliography}

\end{document}